\documentclass{article}

\usepackage{times}  
\usepackage{helvet}  
\usepackage{courier}  
\usepackage[hyphens]{url}  
\usepackage{graphicx} 
\usepackage{natbib}  
\usepackage{caption} 
\DeclareCaptionStyle{ruled}{labelfont=normalfont,labelsep=colon,strut=off} 
\usepackage[algo2e, linesnumbered]{algorithm2e}

\usepackage{xcolor}
\usepackage{newfloat}
\usepackage{listings}
\usepackage[utf8]{inputenc} 
\usepackage[T1]{fontenc}    
\usepackage{url}            
\usepackage{booktabs}       
\usepackage{amsfonts}       
\usepackage{nicefrac}       
\usepackage{microtype}      
\usepackage{subfig}
\usepackage{wrapfig}
\usepackage{lipsum}
\usepackage{amsmath}

\usepackage{caption}
\usepackage{comment}
\usepackage{mathtools}
\usepackage{tabto}
\usepackage{xcolor}
\usepackage{multirow}
\usepackage{array}
\newcolumntype{P}[1]{>{\centering\arraybackslash}p{#1}}
\newcolumntype{M}[1]{>{\centering\arraybackslash}m{#1}}
\usepackage{subfig}
\usepackage{tabularx}
\SetKwInput{KwInput}{Input}
\SetKwInput{KwConstraints}{User Constraints}
\SetKwInput{KwParams}{Parameters}
\SetKwInput{KwOutput}{Output}
\usepackage{hyperref}   
\usepackage{cleveref}
\usepackage{dsfont}

\newcommand{\FRS}{\mathcal{F}}
\newcommand{\BP}{\mathcal{P}}
\DeclareMathOperator{\cov}{cov}

\newcommand{\modnone}{\textit{none}}
\newcommand{\moddrop}{\textit{drop}}
\newcommand{\modrelabel}{\textit{relabel}}
\newcommand{\BPrand}{\textit{random}}
\newcommand{\BPIP}{\textit{IP}}
\newcommand{\mFRS}{\tilde{\mathcal{F}}}

\DeclareMathSymbol{\shortminus}{\mathbin}{AMSa}{"39}

\usepackage[accepted]{mlsys2022}

\mlsystitlerunning{FROTE: Feedback Rule-Driven Oversampling for Editing Models}

\begin{document}

\twocolumn[
\mlsystitle{FROTE: Feedback Rule-Driven Oversampling for Editing Models}




\begin{mlsysauthorlist}
\mlsysauthor{Öznur Alkan}{optum}
\mlsysauthor{Dennis Wei}{ibm}
\mlsysauthor{Massimiliano Mattetti}{mic}
\mlsysauthor{Rahul Nair} {ibm}
\mlsysauthor{Elizabeth M. Daly} {ibm}
\mlsysauthor{Diptikalyan Saha} {ibm}

\end{mlsysauthorlist}

\mlsysaffiliation{ibm}{IBM Research}
\mlsysaffiliation{optum}{Optum}
\mlsysaffiliation{mic}{Microsoft}

\mlsyscorrespondingauthor{Öznur Alkan}{oznur.alkan@optum.com}
\mlsyscorrespondingauthor{Dennis Wei}{dwei@us.ibm.com}
\mlsyscorrespondingauthor{Massimiliano Mattetti}{mmattetti@microsoft.com}
\mlsyscorrespondingauthor{Rahul Nair}{rahul.nair@ie.ibm.com}
\mlsyscorrespondingauthor{Elizabeth M. Daly}{elizabeth.daly@ie.ibm.com}
\mlsyscorrespondingauthor{Diptikalyan Saha}{diptsaha@in.ibm.com}

\mlsyskeywords{Machine Learning, MLSys}

\vskip 0.3in

\begin{abstract}
 Machine learning (ML) models may involve decision boundaries that change over time due to updates to rules and regulations, such as in loan approvals or claims management. However, in such scenarios, it may take time for sufficient training data to accumulate in order to retrain the model to reflect the new decision boundaries.  While work has been done to reinforce existing decision boundaries, very little has been done to cover these scenarios where decision boundaries of the ML models should change in order to reflect new rules. In this paper, we focus on user-provided feedback \emph{rules} as a way to expedite the ML models' update process, and we formally introduce the problem of pre-processing training data to edit an ML model in response to feedback rules such that once the model is retrained on the pre-processed data, its decision boundaries align more closely with the rules. To solve this problem, we propose a novel data augmentation method, the \textit{F}eedback \textit{R}ule-Based \textit{O}versampling \textit{Te}chnique (FROTE). 
 Extensive experiments using different ML models and  real world datasets demonstrate the effectiveness of the method, in particular the benefit of augmentation and the ability to handle many feedback rules.
\end{abstract}
]



\printAffiliationsAndNotice{}  

\section{Introduction}

Machine learning (ML) classifiers are increasingly employed in critical decision-making processes such as loan approvals, credit score assignment \cite{khandani2010consumer}, and claims management \cite{singhuse}. Much focus in the research community has been on improving accuracy of such ML models, evaluated on test data with a similar distribution as the training data. 
However, to deploy such ML models in the real world, one must address problems that arise from the model being inherently governed and limited by the training data. In many applications, domain expert knowledge could be used to improve performance either where data coverage is sparse, or where decision boundaries may have changed over time. Loan approval policies are an example where training data may reflect historical policies but not new policies with shifted decision boundaries. 

Naive options for incorporating expert feedback include manually relabelling historical data and labelling new data. Both are costly in terms of human intervention, and doing the latter alone compromises the accuracy of the deployed model until enough new data is collected. While active learning can reduce the amount of new data needed, the burden may still be too high  \cite{cakmak2010designing, guillory2011simultaneous}, and moreover during deployment, it may not be possible to select which instances to label. Recent work \cite{daly2021aaai} has proposed a more efficient feedback mechanism using rules. This approach uses algorithms for learning decision rules 
\cite{Interpretable_Decision_Sets_KDD2016, ribeiro2018anchors, brcg2018} to provide explanations for arbitrary ML classifiers. The expert's task is then limited to reviewing and modifying a set of classifier predictions and rule-based explanations, resulting in a \emph{feedback rule set} (FRS). \citet{daly2021aaai} propose a post-processing layer 
to account for the feedback rules; however, the feedback is not incorporated into the underlying model.  
\begin{figure*}[t!]
\captionsetup[subfigure]{justification=centering}
\centering
\subfloat{
\includegraphics[width=.3\textwidth,height=12\baselineskip,keepaspectratio]{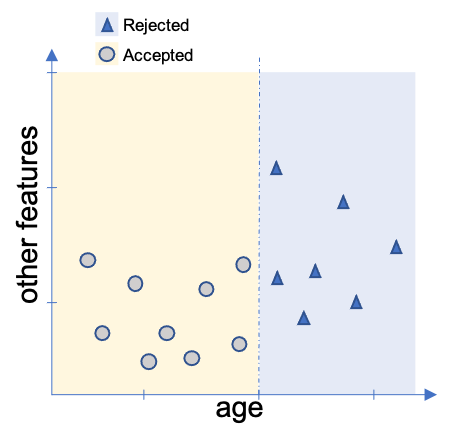}
}
\hfill
\subfloat{
\includegraphics[width=.3\textwidth,height=12\baselineskip,keepaspectratio]{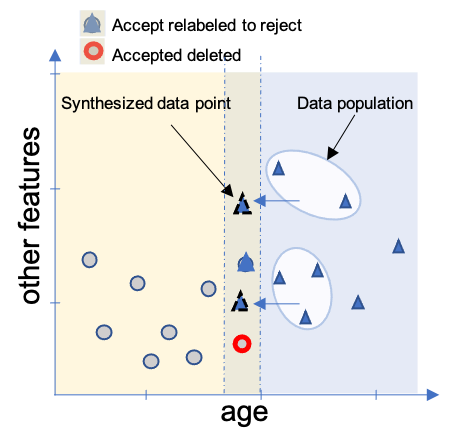}
}
\hfill
\subfloat{
\includegraphics[width=.3\textwidth,height=12\baselineskip,keepaspectratio]{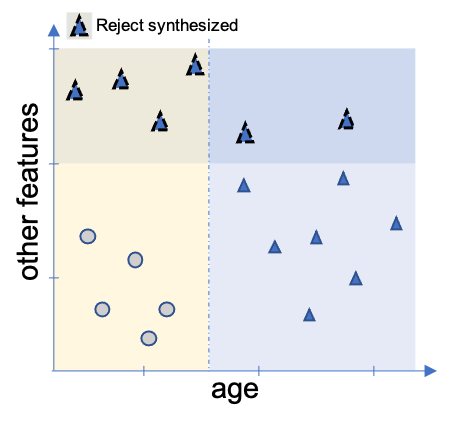}
}

\caption{\textit{Left:} Original classification boundary. \textit{Middle:} FROTE generates synthetic instances to move decision boundary (after relabelling and removing if permitted). \textit{Right:} Generating synthetic instances where existing data is limited.}
\label{fig:mllearning}
\end{figure*}

In this paper, we propose an algorithm called FROTE (\textit{F}eedback \textit{R}ule-Based \textit{O}versampling \textit{Te}chnique) to edit an ML model for tabular data in response to user feedback rules. FROTE thus complements the input transformation method of \cite{daly2021aaai}. 
Given an input dataset, the algorithm first modifies the training data if allowed, and then 
augments it so that re-training the model on the augmented data results in better alignment with the feedback rules. FROTE can thus be used with any classification algorithm that takes training data as input and produces a classifier as output; the algorithm (which could be proprietary) 
is treated as a black box. Unlike \citet{daly2021aaai}, the user feedback is directly encoded in the model.

We use Figure~\ref{fig:mllearning} to be suggestive of a loan approval scenario and to illustrate our solution. Suppose there is a new policy to lower the ages of applicants for whom loans are approved. Rather than crafting rules from scratch, the user relies on the existing ML model and accompanying rule-based explanations to capture relevant dependencies among a potentially large number of features, and only modifies rules that involve age. Given the resulting feedback rule set, the user may wish to 
relabel and remove existing instances as shown in Figure~\ref{fig:mllearning}(b). FROTE then 
generates synthetic instances that reflect both the feedback rules as well as the existing data. Synthetic data generation can address the challenge of insufficient training data in the region to be adjusted, as seen in Figure~\ref{fig:mllearning}(c). 
For data generation, we build upon the SMOTE method \cite{SMOTE_orig} in several ways; 
other methods could also be adapted. 

Our contributions can be summarized as follows: 1) We formulate the problem of editing an ML model by pre-processing a dataset based on user feedback rules. 2) A novel data augmentation-based solution, FROTE, is presented. 3) FROTE is extensively evaluated using different ML models, real-world datasets, and feedback rule set parameters to demonstrate its effectiveness, in particular the benefit of augmentation, improved performance over the state-of-the-art, and the ability to handle many feedback rules.

\section{Related Work}
To the best of our knowledge, the problem studied in this paper is novel in that it differs in at least one of the following aspects from the existing literature: 1) general editing of ML models 2) based on user-specified feedback rules 3) via model-agnostic data augmentation/pre-processing, where the rules can enforce existing boundaries, or introduce new boundaries through changing the dataset. 

Data augmentation/pre-processing has been explored in different problem settings. 
The \textit{class imbalance} problem, which deals with the unequal distribution of classes in training data, 
was tackled in the seminal work of \citet{SMOTE_orig}. 
Their Synthetic Minority Oversampling Technique (SMOTE) randomly selects minority data points as base instances and generates new data points that are convex combinations of the base instances and their $k$ nearest neighbours. 
\citet{han2005borderline} extend SMOTE by synthesizing data points that reinforce existing decision boundaries. Due to its simplicity in the design of the procedure, as well as its robustness, SMOTE has been applied to different type of problems and has proven successful in a variety of applications from several different domains \cite{smoteAnniversary}. 
While we build on SMOTE for data generation, our model editing use case 
differs in going beyond reinforcing existing boundaries to 
adjusting and introducing new ones.
While our contributions build upon these prior works in terms of generating synthetic data instances, our use case is not only to reinforce existing decision boundaries, but also to enable a user both to adjust those decision boundaries and introduce new ones. 

More recently, a more specific use case has gained attention, where data is processed in order to understand and mitigate underlying biases through focusing on fairness. 
Within the \textit{fairness} and \textit{bias mitigation} literature, pre-processing methods such as relabelling and reweighing \cite{calders2009building}, 
data synthesis \cite{AIES_DataAug_Fairness_2020}, and data transformation \cite{optimizedpre} have been proposed.

We argue that the problem we tackle is a more general form of user feedback that can support user concerns through feedback rules, rather than the ones based on only the specified protected features.

Within the \textit{transfer learning} literature, 
\citet{dai2007boosting,eaton2011selective} address a similar problem where test data does not follow the same distribution as training data. They propose an iterative mechanism that re-weights the old data to minimize error observed on the new data. 
In \cite{eaton2011selective}, desJardins and Eaton pursue a similar strategy. 
Neither approach however generates synthetic instances.

In the \textit{generative models} domain, synthetic data generation is used for several tasks. For example, generative adversarial networks (GANs) aim to improve the realism of generated samples until the adversary cannot distinguish real from synthetic data~
\cite{goodfellow2014generative}. In~\citet{tanaka2019data,douzas2018effective,CTGAN}, GANs and conditional GANs with different network architectures 
are used to generate synthetic data to overcome class imbalance as well as privacy issues. In~\cite{douzas2018effective}, a conditional version of GAN (cGAN) is used to generate data for the minority class of various imbalanced datasets. Overall, when comparing the performance of the classifiers on imbalanced data sets that were augmented by the GAN and SMOTE, the former provides better results but with the cost of an higher complexity correlated to the training of the networks. 
Xu et al.~\cite{CTGAN} generate tabular synthetic data using conditional tabular GANs. 
Again these do not support model editing based on rules. 

Incorporating prior knowledge into support vector machines (SVM) was reviewed by \citet{LAUER20081578}. Two forms of prior knowledge were considered: 1. invariances to transformations, to permutations and in domains of input space, 2. knowledge on the unlabelled data, the imbalance of the training set or the quality of the data. 
\citet{SVM_inc_rules_2006} make use 
of knowledge bases of 
rules and virtual support vectors 
to add constraints to the optimization. Different from our solution, these works target only SVM models. Another work from Kapoor et al. 
\citet{kapoor2010interactive} support user influence over ML algorithms by manipulating confusion matrices, where the user is allowed to manipulate the initial confusion matrix over the different classes. 

Leveraging expert rules has been explored in the \textit{assisted labelling} literature. Snorkel \cite{ratner2017snorkel} takes a weak supervision approach to labelling training data by bringing together label predictions from different sources, including labelling functions that can be expert-provided patterns. The labelling sources include labelling functions which can be expert provided patterns and heuristics to predict labels.  A generative model is built to estimate the accuracy and correlations of the different labelling sources and produces probabilistic training data where each data point has a probabilities distribution over all the labels and then can be used to train a model. 
\citet{awasthi2020learning} consider hybrid supervision from labelled instances as well as rules that generalize them. 
The assisted labelling problem is different from ours in that they seek to label unlabelled data whereas we already have a model trained on a labelled dataset and wish to edit the model, without negatively impacting accuracy for data unaffected by the rules. In addition, in assisted labelling, experts have to devise rules from scratch whereas in model editing, they may only have to modify rules that capture what the model has already learned. They provide a solution where labels are unavailable or noisy and seek to label unlabelled data. Our goal is somewhat different, where we assume the presence of a dataset and a model that may be considered trusted and validated but the user wants to make some adjustments or edits without negatively impacting the model accuracy for unaffected data which should remain unchanged.  


The most closely related work by \citet{daly2021aaai} addresses user feedback rules, but not by editing the ML model. Instead, transformations that map between the original and feedback rules are obtained to yield a post-processing layer called Overlay. When a new data point arrives for prediction, Overlay checks to see if a feedback rule corresponds to the data point and if so, applies the transformation, returning the prediction of the transformed data point. 
While Overlay enables immediate 
changes to an ML system by applying the above transformations to the input, 
without retraining the model, \citet{daly2021aaai} note that it is a ``patch''. As more feedback rules and their corresponding patches are produced, the overall system consisting of the ML model and these patches 
may become overly complex and difficult to maintain. It is not difficult to imagine that even a single expert could generate a large number of feedback rules. 
Additionally, experiments by \citet{daly2021aaai} suggest and our experiments confirm (Table~\ref{table:overlay}) that one limitation of Overlay occurs when a feedback rule differs too significantly from the 
underlying model, a limitation that FROTE overcomes. Moreover, in applications such as finance or spam detection, Overlay's transformations may incur additional undesirable latency. 
For the reasons above, once short-term patches have been applied, 
it may be preferable to directly incorporate user feedback into the model, which is the problem that FROTE solves.

\section{Preliminaries}

As discussed in the Introduction, the premise of this work is that 1) the distribution of future data (i.e.~test data) is different from that of training data, due for example to a policy change or to the training data not being representative, and 2) a domain expert understands the nature of the change and communicates that through a set $\FRS$ of \emph{feedback rules}, i.e. a \emph{feedback rule set} (FRS). To establish notation, let $\mathbf{x} \in \mathcal{X} \subset \mathbb{R}^d$ denote a set of attributes for decision-making, and $y \in \mathcal{Y} = \{c_1, c_2, \dots, c_l\}$ denote a class label. The existing training data is a set of $n$ instances $(\mathbf{x}_i, y_i)$, $i = 1,\dots,n$, assumed to be drawn i.i.d.~from a joint distribution $p_{X,Y}$.

\subsection{Feedback Rules}

We consider a generalization of decision rules beyond recent works \cite{Interpretable_Decision_Sets_KDD2016,molnar2019} to allow feedback rules that are \emph{probabilistic}. A feedback rule $R = (s, \pi)$ is thus a statement of the form IF \textit{the clause $s$ is true} THEN \textit{the class label $Y$ is distributed according to $\pi$}. These are discussed in turn below.

\paragraph{Clauses and coverage.} A clause is a conjunction of one or more predicates (also referred to as conditions) of the form (attribute, operator, value). In our solution, the operators allowed for categorical attributes are \{=, $\ne$\}, and for numeric attributes are \{=, $>$, $\ge$, $<$, $\le$\}. An example of a clause with three predicates is \textit{age $<$ 29 AND marital-status = `single' AND income $>$ 150K}. We say that $\mathbf{x} \in \mathcal{X}$ \textit{satisfies} a clause $s$, and reciprocally, a rule $(s, \pi)$ \emph{covers} $\mathbf{x}$, if all the predicates in $s$ are \textit{true} when evaluated on $\mathbf{x}$. Given a dataset $D$, \emph{coverage} of a rule $(s, \pi)$ and an FRS $\FRS = \{(s_r, \pi_r)\}_{r=1}^m$ of $m$ feedback rules are defined as follows:

\begin{align}
    \cov(s, {D}) &= \{(\mathbf{x}, y) \in {D} : \mathbf{x} \text{ satisfies } s\},\\ \cov(\FRS,{D}) &= \bigcup_{r=1}^m \cov(s_r,{D}).
\end{align}
Note that coverage involves only clauses $s$ and attributes $\mathbf{x}$. If $D$ is omitted as in $\cov(s)$, then it is understood to be the entire domain $\mathcal{X}$. 

The reason for using logical clauses as above is that they semantically resemble natural language and the way humans think~\cite{interprrules2017, interpretablerules2015, molnar2019}. Therefore it can be more natural for users to provide feedback in the form of a rule, either of their own creation or by modifying an algorithm-provided rule-based explanation. This does require the rule's conditions to be built from intelligible features and favours smaller numbers of conditions and rules~\cite{Interpretable_Decision_Sets_KDD2016}. 

\paragraph{Label distribution.} Given a feedback rule $(s, \pi)$ and $\mathbf{x} \in \cov(s)$, we assume that the class label is distributed as $Y \sim \pi$. 
We will mostly work with the \emph{deterministic} case where $\pi$ is the Kronecker delta distribution for a class $c$, i.e., $Y = c$ with probability $1$. This is the easiest case for a human expert, who only has to specify the class $c$. However, allowing probabilistic rules is useful for at least two reasons: 1) accommodating conflicts between rules (discussed next), and 2) allowing uncertainty in rules and providing robustness against over-confident 
rules.

\paragraph{Rule conflicts.} When feedback from multiple experts is to be considered, the possibility of conflicts should be taken into account due to contradictory opinions. Two rules $(s_1, \pi_1)$, $(s_2, \pi_2)$ are conflicting if their coverages intersect, $\cov(s_1) \cap \cov(s_2) \neq \emptyset$, and $\pi_1 \neq \pi_2$. 
We assume that all such conflicts are resolved, for example through one of the following options:
\begin{enumerate}
    \item Removal of the intersection, i.e., clause $s_1$ is changed to $s_1$ \textit{AND NOT} $s_2$, and $s_2$ to $s_2$ \textit{AND NOT} $s_1$.
    \item Creation of a new rule for the intersection with a mixture of the distributions, e.g.~$(\pi_1 + \pi_2) / 2$ or a more general weighting. The intersection is then excluded from the two original rules as in option 1.
    \item If the two rules are provided by different experts, asking them to come to a consensus.
\end{enumerate}
We assume that the final FRS is \emph{conflict-free} through repeated application of the above operations for conflict resolution.

\subsection{Problem Formalization}

We are given 1) a conflict-free feedback rule set $\FRS$, 2) an initial training dataset $D$, and 3) a classification \emph{algorithm} $A$ that, given a dataset $D$, trains a classification model $M_D$. The task is to create a dataset ${\hat D}$ by 
augmenting $D$ such that when the model is retrained on ${\hat D}$ using $A$ to yield ${M}_{\hat D}$, the objective function in 
\eqref{eq:objf}  is minimized. To define the objective function, let $L_1, L_2: \mathcal{Y} \times \mathcal{Y} \mapsto \mathbb{R}$ be two loss functions that compare two labels. We also assume for ease of exposition that the rule coverage sets are disjoint, which can be achieved by 1) resolving conflicts as described above, and 2) merging rules that overlap but do not conflict. 
Then the objective function can be written as;
\begin{align}
    &J\bigl(M_{\hat D}, \FRS\bigr) = 
\sum_{(s_r,\pi_r)\in\FRS} \Pr(X \in \cov(s_r)) \nonumber\\ 
&{} \times \mathbb{E}_{X\sim p_X, Y\sim\pi_r}\left[ L_1\bigl(M_{\hat D}(X), Y\bigr) \:|\: X \in \cov(s_r) \right] \nonumber\\
&{} + \Pr(X \notin \cov(\FRS)) \mathbb{E}_{X,Y\sim p_{X,Y}}\left[ L_2\bigl(M_{\hat D}(X), Y\bigr) \:|\: X \notin \cov(\FRS) \right]. \label{eq:objf}
\end{align}

The summation in \eqref{eq:objf} applies to instances in the coverage of the FRS and evaluates the retrained model's predictions $M_{\hat{D}}(X)$ against labels $Y$ distributed according to each feedback rule's $\pi_r$. We refer to the complement of this term (i.e.~$1$ minus it) as \emph{model-rule agreement} (MRA). The motivation for the name MRA comes from the case where $L_1$ is the $0$-$1$ loss. Then the expectation of $1 - L_1(M_{\hat{D}}(X), Y)$ is the probability of agreement between $M_{\hat{D}}(X)$ and $Y$. 

The last term in \eqref{eq:objf} applies to instances outside $\cov(\FRS)$ and evaluates the predictions against labels following the original distribution $p_{X,Y}$. We refer to this term as outside-coverage performance.

\section{Proposed Solution}
\label{sec:solution}
Given an input dataset $D$, the goal of our proposed solution FROTE is to produce an augmented dataset $\hat{D}$ so that retraining the model on $\hat{D}$ minimizes the loss function defined in equation~\eqref{eq:objf}. 
%
%
%
%
The initial dataset $D$ could be the one used to train the original model, or it could be a modified version of this dataset. We show in the Experiments section and supplement that FROTE works with different types of initial datasets. The steps of FROTE are given in Algorithm~\ref{alg:FROTE}. 

\textbf{Base instance selection.}  The adaptation of SMOTE used by FROTE requires a set of \emph{base instances} chosen from the original dataset. These provide the basis for augmentation to ensure that generated instances are similar to original instances. Base instance selection occurs in two steps: pre-selection of a \emph{base population} (BP), denoted $\BP$, before the main augmentation loop (line 4), and selection of subsets of the BP, denoted $\mathcal{B}$, within the loop (line 7). These are described in the Base Instance Selection subsection. 

\textbf{Augmentation loop.} In each iteration of FROTE, base instances are selected from the BP (line 7) and corresponding synthetic instances are generated (line 8) as described in the Synthetic Instance Generation subsection. 
A temporary dataset $D'$ is created (line 9) by combining these synthetic instances with $\hat D$, the current active dataset. 
The model is retrained on $D'$ (line 10) and the loss function $J$ is evaluated (line 11). If the loss decreases (lines 12-15), $D'$ becomes the current active dataset 
$\hat D$. Otherwise, the generated instances are discarded and $\hat D$ is unchanged. This augmentation loop proceeds until one of the termination criteria is met: 
1. the \textit{oversampling quota} (controlled by oversampling fraction $q$) is used up, or 2. the \textit{iteration limit} $\tau$ is exceeded. 

\textbf{User Constraints.} We regard $\tau$ and $q$ as constraints determined by user preferences: $\tau$ is the number of times the user is willing to run training algorithm $A$, and $q$ is the allowed amount of augmentation relative to the initial dataset. Given $\tau$ and $q$, the number of generated instances per iteration is set to $q \lvert D \rvert / \tau$ (line 1) to uniformly distribute the quota. 

\begin{algorithm}[t]
\SetKwFor{RepTimes}{repeat}{times}{end}

\small
\DontPrintSemicolon
\KwInput{
  input dataset $D$, ML algorithm $A$, feedback rule set $\FRS$}
\KwConstraints{iteration limit $\tau$,  oversampling fraction $\mathit{q}$} 
\KwOutput{output dataset $\hat D$}
\BlankLine
$\eta \gets q \lvert D \rvert / \tau$, $\hat D \gets D$ \;

$M_{\hat D} \gets$ apply training algorithm $A$ to $\hat D$ \;

$\hat j \gets \hat{J}_{\hat{D}}(M_{\hat D}, \FRS)$\;

$\BP \gets \mathrm{PreSelectBP}(\hat D, \FRS)$\;

$\mathit{i}, \mathit{N} \gets 0$ \;

\While{$i < \tau$ and $\mathit{N} \le \mathit{q} \times |D|$} {
 $\mathcal{B} \gets \mathrm{SelectBaseInstances}(\BP, \eta)$ \;
 
 $\mathcal{S} \gets \mathrm{Generate}(\mathcal{B})$ \;
 
 ${D}' \gets \hat{D} \cup \mathcal{S}$\;
 
 $M_{D'} \gets$ apply training algorithm $A$ to $D'$ \;
 
 $j'$ $\gets \hat{J}_{\hat{D}}(M_{D'}, \FRS)$\;
 
 \If{$j'$ $ <\hat j$}
 {
    $\hat D \gets D'$, $\mathit{N} \gets \mathit{N} + |S|$\;
    
    $\hat j \gets j'$\;
    
    $\BP \gets \mathrm{PreSelectBP}(\hat D, \FRS)$, \;
    
 }
 $\mathit{i} \gets \mathit{i} + 1$
}

\caption{FROTE}
\label{alg:FROTE}
\end{algorithm}
\subsection{Base Instance Selection}
\label{sec:solution:BP}

Whereas SMOTE randomly selects data points from the minority class as the base population, our problem is more challenging as it is driven by the loss $J$ in \eqref{eq:objf} and the
ideal selection of base instances would maximally decrease this loss. 
Referring to Algorithm~\ref{alg:FROTE}, we denote by $\mathcal{B}$ the set of selected base instances, $\mathcal{S} = \mathrm{Generate}(\mathcal{B})$ the synthetic instances generated from $\mathcal{B}$, and $A(D')$ the model obtained from the temporary dataset $D' = \hat{D} \cup \mathcal{S}$. Then the goal is to choose $\mathcal{B}$ to minimize
\begin{equation}\label{eqn:J(B)}
    J(M_{D'}, \FRS) = J\bigl(A\bigl(\hat{D} \cup \mathrm{Generate}(\mathcal{B})\bigr), \FRS\bigr).
\end{equation}

There are multiple challenges in minimizing \eqref{eqn:J(B)}: 1) Choosing $\mathcal{B}$ is a combinatorial subset selection problem. The size of the subset $\lvert\mathcal{B}\rvert = \eta$ may be large (e.g.~$100$), and the size of the BP $\BP$ is larger still. 2) The training algorithm $A$ is a black box. Furthermore, it may be expensive to run to evaluate \eqref{eqn:J(B)}. 3) The expectations in $J$ must be approximated with empirical averages. We address this by using the current active dataset $\hat{D}$, replacing $J$ with the empirical approximation $\hat{J}_{\hat{D}}$ over $\hat{D}$ (lines 3, 11). As a consequence however, even evaluating \eqref{eqn:J(B)} for all singleton $\mathcal{B}$, e.g.~all instances in $\BP$, would incur complexity of at least $O(\lvert\BP\rvert\lvert\hat{D}\rvert)$. This implies that even a greedy selection algorithm, which would evaluate $O(\eta\lvert\BP\rvert)$ subsets, would have cubic complexity $O(q\lvert D\rvert^3 / \tau)$ assuming $\eta = q \lvert D \rvert / \tau$ and $\lvert\BP\rvert \propto \lvert D \rvert$.

Herein we take a simple approach to base instance selection, consisting of 1) pre-selecting a BP to focus only on the coverage set $\cov(\FRS, D)$, 2) selecting subsets \emph{randomly}, and 3) exploring more informed strategies that maintain low computational complexity.

\textbf{Base population pre-selection (line 4).} Motivated by the MRA term in equation~(\ref{eq:objf}), we restrict the BP to the coverage $\cov(\FRS, D)$. In our implementation, we maintain \emph{per-rule} BPs, i.e., $\BP[r]$ for $R_r \in \FRS$, and accordingly initialize $\BP[r] = \cov(s_r, D)$. However, rules may have little or no coverage in the original dataset $D$, and the method described in the Synthetic Instance Generation subsection requires coverage of at least $k+1$. To handle this scenario, FROTE uses \textit{rule relaxation} to obtain a \emph{maximal} partial rule set, denoted as $\mFRS$. During augmentation, an instance is selected to be part of the base population if it is \emph{strongly covered}, i.e. the instance matches a rule within $\FRS$ exactly, or if it is \emph{weakly covered}, i.e. the instance only matches a rule partially. The latter case is designed to handle a relaxed case when a rule in $\FRS$ has zero support. In this case, we determine the maximal partial rule, a version of the rule with the minimal condition deletion that gives the maximum support. In other words, we tried to find out the minimum change we can make to the rule to give us the largest non-zero support. Since the number of conditions within each rule set is low, such a maximal partial rule can be determined by a breath-first search exhaustively by first removing one condition and then two and so on.

\paragraph{Base population pre-selection.} 
\SetKwFor{RepTimes}{repeat}{times}{end}
\begin{algorithm}
\small
\KwInput{input dataset $D$, \\feedback rule set $\FRS$,\\
 number of nearest neighbours $\mathit{k}$
}
\KwOutput{initial base population $BP$}
\SetAlgoLined
\setcounter{AlgoLine}{0}
\LinesNumbered
\DontPrintSemicolon
$L \gets k+1$ \;

$BP \gets \emptyset$ \;

\For{each rule $R$ in $\FRS$}
{
 \If{$cov(R,D)$ $ <L$}
 {
    max\_sup $\gets 0$ \;
    
    max\_cond\_R $\gets nil$\;
    
    \While {max\_sup $< L$}{
    
        \For{each condition $c$ in $R^s$} {
    
            $R' \gets R$ \;
            
            remove condition $c$ from $R'^s$ \;
            
            \If{$R'^s$ is empty} {
                max\_sup $\gets |D|$\;
                
                max\_cond\_R $\gets R'$\;
                
            }
            \Else{
                \If{$cov(R',D)$ $ >$ max\_sup} {
                    max\_sup $\gets cov(R',D)$\;
                    
                    max\_cond\_R $\gets R'$\;
                }
            }
            $R \gets$ max\_cond\_R
        }
    }
 }
 $BP \gets BP \cup cov(R,D)$
}
\caption{PreSelectBP}
\label{alg:PreSelectBP}
\end{algorithm}

Base population pre-selection procedure \textit{PreSelectBP} is outlined in Algorithm~\ref{alg:PreSelectBP}. For each rule in the feedback rule set $\FRS$, FROTE requires coverage of at least $k+1$ to generate synthetic instances, where $k$ represents the number of nearest neighbors. Therefore, conditions of a feedback rule $R$ are relaxed if the coverage of $R$ is less than $k+1$ (lines 7-18). During rule relaxation, the goal is to remove minimum number of conditions from $R^s$ that will result in a maximum rule coverage. To achieve this, \textit{PreSelectBP} performs a breadth first search on a tree of |$R^s$| levels, where at each level the nodes are the remaining conditions in $R^s$. At each level, \textit{PreSelectBP} chooses a condition whose removal results in maximum coverage in comparison with other conditions that exist at that level (lines 8-18). The procedure returns the union of the instances within the coverage of the relaxed feedback rules.




\textbf{Random subset selection (line 7).} The simplest choice for selecting base instances is to randomly select $\eta$ instances from the BP, motivated in part by~\citet{SMOTE_orig}. We refer to this strategy as \BPrand{} in the paper. More specifically, base instances are selected on a per-rule basis as detailed in the supplement. Despite its simplicity, we find during the experiments that \BPrand{} appears to work well empirically.  

\textbf{Subset selection via integer programming (line 7).} We also consider an integer programming (IP) approach, referred to as \BPIP. Unlike \BPrand, \BPIP{} takes into account the current ML model $M_{\hat{D}}$ in seeking to generate synthetic instances that have a greater effect on the objective $J$. The model is accounted for using \textit{borderline} instances, which are data points that lie close to the decision boundaries of the model and thus have more impact~\cite{han2005borderline}. 

To quantify the value of different base instances, we associate a weight $w_i$ with each base instance $i$ in the BP $\BP$. Weights are pre-computed using a similar strategy followed in~\citet{han2005borderline}, where instances are classified as \textit{noisy}, \textit{safe}, or \textit{borderline} based on the number of nearest neighbours with the same and different class labels, and the highest weight is assigned to borderline instances (see supplement for details).  

Let $z_i$ be a binary variable such that $z_i=1$ if the $i$-th instance in the BP $\BP$ is selected, and $z_i = 0$ otherwise. Given $\BP$, we define a matrix $\mathbf{A}$ with entries $a_{ji}$ and dimensions $m \times p$, where $m$ represents the number of rules and $p = \lvert \BP \rvert$, 
such that $a_{ji}=1$ if instance $i$ is covered by feedback rule $j$ and $a_{ji}=0$ otherwise. Then the problem of selecting base instances can be stated as the following IP:
\begin{equation}
\max_{z \in \{0,1\}^p} \sum_{i \in \BP} w_i z_{i}, 
\textrm{s.t.}, k+1 \le \sum_{i \in \BP} a_{ji} z_i \le \frac{\eta}{m}, j=1,\dots,m.\label{eq:IP}
\end{equation}
The objective is to maximize the weighted selection of base instances subject 
to lower and upper bounds on the number of instances selected for each rule. Since the data augmentation step described in the next section seeks $k$ neighbours, the lower bound is set to $k + 1$. This also preserves the per-rule diversity in the BP. The upper bound is the number of instances to generate divided by the number of rules. Non-uniform allocations of instances to rules are also possible. 

Despite \eqref{eq:IP} being an IP, in practice it can be solved quickly as linear relaxations directly provide integral optimal solutions in most cases. 
Furthermore, $\BPIP{}$ avoids any evaluation of the objective function \eqref{eqn:J(B)} in selecting base instances. In the supplement, we also discuss an approach that simplifies the evaluation of \eqref{eqn:J(B)} by using online learning in place of the more expensive black-box algorithm $A$.

\subsection{Synthetic Instance Generation}
\label{sec:oversampling}
Motivated from SMOTE and its extension to categorical attributes, SMOTE-NC~\cite{SMOTE_orig}, we design a methodology to generate synthetic instances (line 8 of Algorithm 1) for each selected base instance in line 7. 
SMOTE generates synthetic instances that lie between a base instance and one of its $k$ nearest neighbours, selected at random. 
For numerical attributes, the generated value is distributed uniformly on the line segment between the base instance and the neighbour. 
For categorical attributes, the value is the majority value among the neighbours. Following the recommendation of \citet{SMOTE_orig,han2005borderline}, we set the number of neighbours $k = 5$.

FROTE's generation method differs from SMOTE in the following ways: First, nearest neighbours are found without the constraint that they have the same class label as the base instance, but with the constraint that they satisfy the same feedback rule (possibly relaxed). Second, we require that the generated instance satisfies the conditions of the original, \emph{unrelaxed} rule. This happens automatically if the rule was not relaxed, but if it was, then special logic is needed as described in the supplement. 
Third, the class label for the generated instance is sampled from the distribution $\pi$ of the rule (or simply assigned if the rule is deterministic) rather than being equal to the label of the base instance.

\begin{figure*} [th]
    \centering

    \centering
        \includegraphics[width=0.8\textwidth]{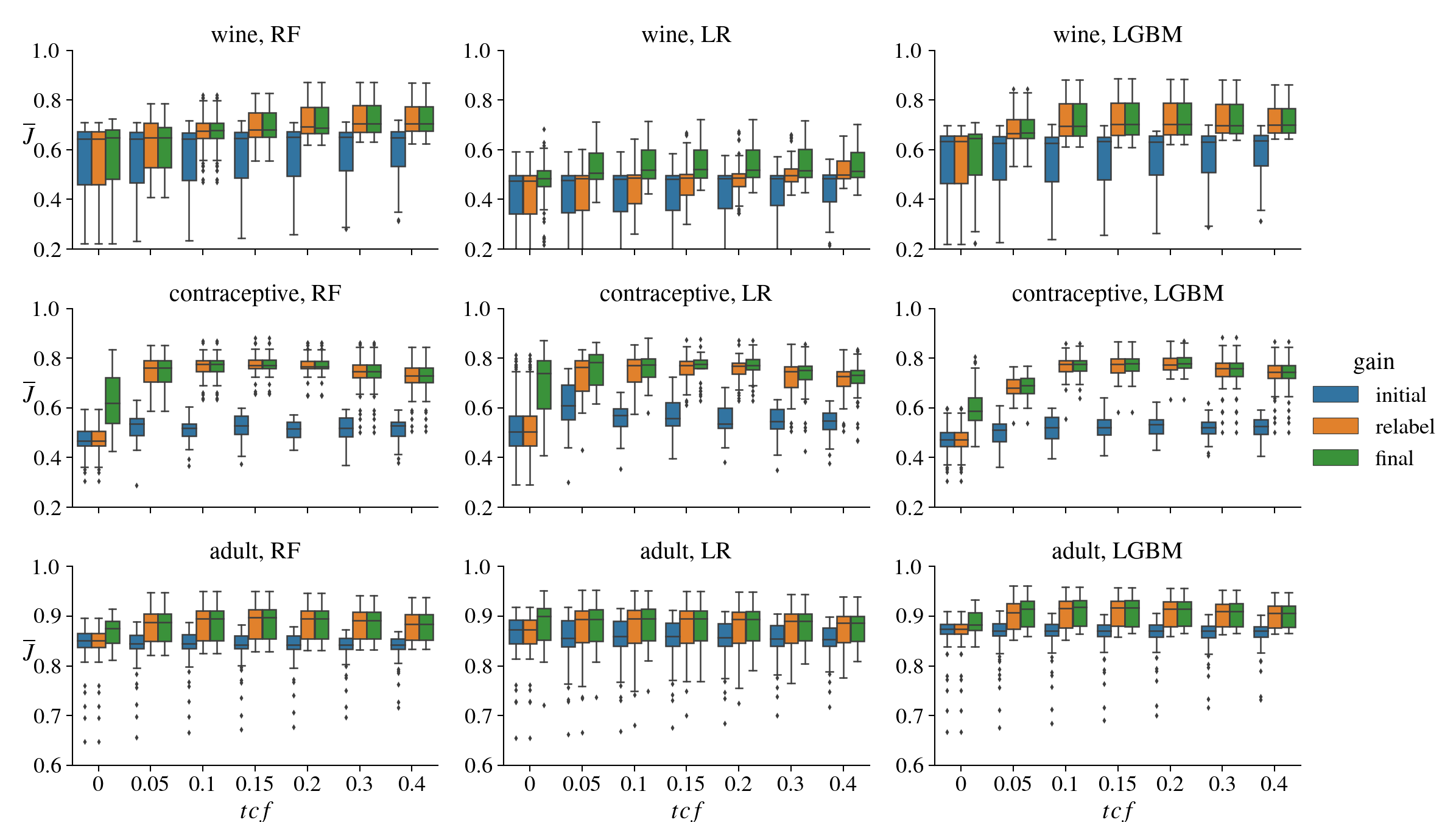}
     \caption{Experiments with models trained on initial training dataset (\textit{initial}), after relabelling (\modrelabel{}), and after FROTE completes augmentation (\textit{final}). 
     $\BPrand$ selection strategy is used. Standard box plot shows interquartile range (IQR) and whiskers show $1.5\times IQR$  based on 30 draws for each of $\lvert\FRS\rvert \in \{1,3,5\}$. Results with other datasets included in Section B. }
    \label{fig:modperf}
\end{figure*}

\section{Experimental Evaluation}
\label{sec:expt}
\subsection{Experimental Setup}
\textbf{Datasets, ML Models, Feedback Rules}

\begin{table}[t]
    \small
    \centering
    \caption{Properties of the datasets used during the experiments. \#Ins, \#Labels, and \#Feat. stands for the number of instances, number of class labels and number of features (numeric/nominal) of the datasets, respectively.}
    \vspace*{2mm}
    \begin{tabular}{|p{2.8cm}| p{1cm}| p{1.2cm}| p{1cm}| p{4cm}}
        \hline
        \textbf{Dataset} & \textbf{\#Ins.}    & \textbf{\#Feat.}    & \textbf{\#Labels }\\ \hline
        Adult & 45222 & 12(4/8) & 2 \\
        Breast Cancer & 569 & 32(32/-) & 2 \\
        Nursery & 12958 & 8(-/8) & 4 \\
        Wine Quality (white)& 4898 & 11(11/-) & 7 \\
        Mushroom & 8124 & 21(-/21) & 2\\
        Contraceptive & 1473 & 9(2/7) & 3  \\
        Car & 1728 & 6(-/6) & 4  \\
        Splice & 3190 & 60(-/60) & 3 \\\hline
    \end{tabular}
    \label{tab:datasets}
\end{table}

To evaluate the effectiveness of FROTE, we experimented with eight real-world benchmark datasets from UCI\footnote{https://archive.ics.uci.edu/ml/datasets.php}, properties of which are provided in Table~\ref{tab:datasets}. 
To generate realistic feedback rules, we follow the process mentioned in the Introduction by leveraging \textit{Boolean Rules via Column Generation (BRCG)} algorithm~\citet{brcg2018} to obtain a rule set explanation for an initial ML model, and then artificially 
perturbing these rules to simulate users providing feedback that deviates from the model's predictions. For each rule extracted from~\citet{brcg2018}, we performed the following three perturbations until we generate $100$ rules for each dataset with coverage satisfying $0.05 \le \lvert\cov(s,D)\rvert/\lvert D\rvert < 0.25$: For each rule extracted, 1. A predicate is randomly selected from the rule's clause and the operator is \textit{reversed}. For instance, if the operator is $\ne$, it is changed to $=$, and similarly if the operator is $\leq$, it is changed to $\geq$, respectively. 2. Value of the selected predicate is updated based on its values in the training dataset. For instance, for categorical attributes, any randomly selected value other than the value of the current predicate is picked and assigned. Similarly for the numerical attributes, a value within the range of the minimum and the maximum values of that attribute observed in the training dataset is assigned. 3. An existing condition from any other rule is randomly picked and added to the rule's conditions. 
We generated $100$ feedback rules in this manner for each dataset,
where each generated rule has coverage satisfying $0.05 \le \lvert\cov(s,D)\rvert/\lvert D\rvert < 0.25$. Rules are deterministic except for the probabilistic rules experiment in Section B.

\textbf{Classification models.} We used three classification algorithms: scikit-learn's Random Forest 
(RF) and Logistic Regression 
(LR), and LightGBM
(LGBM) \cite{ke2017lightgbm}. Default parameter settings are used except for \texttt{max\_iter} $=500$ for LR and \texttt{max\_depth} $=3$ for RF. 
For finding nearest neighbours in FROTE, scikit-learn's Nearest Neighbors
\cite{scikit-learn} algorithm with \texttt{algorithm=ball\_tree} is utilized. 

\textbf{FRS selection and train-test splitting.}
We experimented with FRS sizes $\lvert\FRS\rvert\in \{1,3,5, 8, 10, 15, 20\}$, and for each run, we randomly draw this many rules from the pools of $100$ generated as described above. We used the following mechanism to vary the level of support of the FRS in the initial training data. 
For each dataset $D$ and FRS $\FRS$, $D$ is partitioned into \textit{coverage} ($\cov(\FRS,D)$) and \textit{outside-coverage} ($D-\cov(\FRS,D)$) sets. $D-\cov(\FRS,D)$ is randomly partitioned on a $(80\%-20\%)$ basis into training and test. 
For the coverage set $\cov(\FRS,D)$, we vary the \textit{training coverage fraction} ($tcf$), i.e.~the fraction of the coverage set included in the training set. That is, $tcf \times \lvert\cov(\FRS,D)\rvert$ randomly selected instances are added to the training partition of $D-\cov(\FRS,D)$, and the remainder to the test partition of $D-\cov(\FRS,D)$. We experimented with $tcf\in \{0, 0.05, 0.1, 0.15, 0.2, 0.3, 0.4\}$. $tcf=0$ tests the scenario where the FRS has no coverage in the initial training set, for example when a new rule emerges.  

We perform $30$ to $50$ runs as described in the previous paragraph for each experimental setting, depending on the size of the dataset. This method of randomly drawing a new rule set and train-test split for each run increases the variability of rule sets tested (and their impact on the results) compared to fixing a rule set and performing cross-validation with it. All algorithm variations are compared using the same rule sets and splits. 

\textbf{Metrics.} FROTE uses only the training dataset for augmentation and all evaluation results are reported on the held-out test set.  
We report values of the complement of $J$, $\overline{J}=1-J$, where $\overline{J}$ is a weighted average as in \eqref{eq:objf}, weighted by rule coverage probabilities $\Pr(\cov(s_r, D))$ in the test set, the first term is the MRA discussed previously (with $L_1$ as $0$-$1$ loss), 
and the last term is $F_1$ score to evaluate model performance on the outside-coverage population. 
In running FROTE however, we simply use a $0.5$-$0.5$ weighting between MRA and $F_1$ score in evaluating $\hat{J}_{\hat{D}}$. This is because the test set coverage probabilities are not known to FROTE and may not be equal to the training set probabilities.


\textbf{Input dataset choices.} We experiment with three choices of input dataset $D$ to FROTE. In addition to 1) taking the training dataset as it is (denoted \modnone{} for no modification), instances in $\cov(\FRS, D)$ that do not have the same class label as the feedback rules covering those instances may be 2) relabelled to agree with the covering rules (\modrelabel{}) or 3) dropped (\moddrop{}). 
\modrelabel{} is used in all experiments except for the one that evaluates input dataset choices. It is important to note that \modrelabel{} and \moddrop{} may not be possible if the user is reluctant to make changes to the existing dataset for various data integrity reasons.


\textbf{Configuration.} The number of instances generated per iteration ($\eta$) is set to $200$ for Adult dataset, $50$ for Nursery, Mushroom, Splice, and Wine datasets, and $20$ for Car, Contraceptive and Breast Cancer datasets. $\mathit{\tau=200}$ is used as the iteration limit for all the experiments. We used $k=5$ and $q=0.5$ for all the experiments except the ones we evaluated the effect of these two parameters. All experiments were limited to $24$ hours  and runs that exceed this time limit were terminated. We ran all experiments on a 2.6GHz CPU with 20GB of RAM and they were run deterministically with consistent random number generator seed (42).

\subsection{Results and Discussion}
\label{sec:expt:results}

\textbf{Benefit of augmentation.} 
In Figure \ref{fig:modperf}, we compare the test set $\overline{J}$ values obtained from models trained on 1) the initial training dataset, 2) after relabelling based on the FRS (\modrelabel{}), and 3) after FROTE completes augmentation. The comparison is shown for the three ML models, a range of training coverages, and three of the datasets with the remainder in Section B. 
Even after relabelling, FROTE's augmentation improves $\overline{J}$ for all models and datasets compared to relabelling alone (\textit{final} vs.~\textit{relabel}). This finding is further supported by similar plots in  Section B of \emph{differences} in $\overline{J}$ between \textit{final} and \textit{relabel}, the vast majority of which are positive. Not surprisingly, the same conclusion holds more strongly for the \moddrop{} and \modnone{} options (see Section B).

Two trends are evident from Figure~\ref{fig:modperf}. First, the improvement over \modrelabel{} is larger for smaller $tcf$, and notably for the difficult case of $tcf = 0$ in which the initial training dataset has no coverage of the FRS. This shows that \modrelabel{} is not sufficient and there is a greater need for augmentation when $tcf$ is low. Second, the improvement is larger for LR, which indicates that linear models may require more data to push decision boundaries.

\textbf{Comparison with the existing work.} To the best of our knowledge, the closest work to ours is Overlay~\cite{daly2021aaai}, which includes two approaches, \textit{Soft Constraints} and \textit{Hard Constraints}. The former treats the user feedback as a soft constraint and uses the prediction on the transformed instance, and the latter considers the feedback as a hard constraint and uses the feedback rules' prediction for all applicable instances. A similar setting as in the previous experiments is used for this comparison. For each run with a dataset, 3 rules are randomly selected and provided as the Full Knowledge Rule Set (FKRS)~\cite{daly2021aaai} for Overlay, and as the FRS for FROTE. For each rule set, $50\%$ of the coverage population is included in the training data and rest in the test data. Similarly, for the outside-coverage population, a $50\%-50\%$ split is performed. The model is trained on the training dataset, and FROTE, \textit{Soft Constraints} and \textit{Hard Constraints} are evaluated on the held-out test set. Overlay is presented for binary classification problem and the experiments reported in~\cite{daly2021aaai} are performed using binary datasets. Therefore we experimented with only the 3 binary datasets (out of 8), and results are displayed in Table~\ref{table:overlay} (Results with the adult dataset together with separate MRA and F-Scores are in Section B.) We observe that FROTE performs significantly better than both approaches of Overlay for all datasets. The performance of \textit{Soft Constraints} and \textit{Hard Constraints} differs greatly, which suggests the user feedback rules are too divergent from the decision boundaries of the initial ML model for Overlay to perform well, in line with the findings of \citet{daly2021aaai}. This demonstrates that our solution for integrating user feedback into models through pre-processing achieves a better performance in comparison with a state-of-art post-processing approach.

\begin{table}[t]
    \small
    \centering
        \caption{Comparison with Overlay-Soft (soft constraints) and Overlay-Hard (hard constraints) of \citet{daly2021aaai} on BreastCancer and Mushroom datasets. Means and standard deviations computed from 50 runs.}
        \vspace*{2mm}
    \begin{tabular}{@{}p{0.9cm}p{0.7cm}p{1.75cm}p{1.78cm}p{1.75cm}@{}}
        \toprule
        Dataset & Model & \multicolumn{3}{c}{$\Delta \overline{J}$}\\
        \cmidrule(lr){3-5}
         & & Overlay-Soft & Overlay-Hard & FROTE\\
        \midrule
        B.Cancer & LR & $\mathllap\shortminus0.008\pm0.045$ & $\mathllap\shortminus0.237\pm0.212$ & $0.030\pm0.008$\\
        & RF & $0.001\pm0.003$ & $\mathllap\shortminus0.215\pm0.204$ & $0.041\pm0.018$\\
        & LGBM & $0.006\pm0.011$ & $\mathllap\shortminus0.207\pm0.180$ & $0.033\pm0.015$\\\hline
        Mushr. & LR & $0.001\pm0.004$ & $\mathllap\shortminus0.158\pm0.213$ & $0.014\pm0.015$\\
        & RF & $0.001\pm0.004$ & $\mathllap\shortminus0.153\pm0.208$ & $0.009\pm0.008$\\
        & LGBM & $\mathllap\shortminus0.017\pm0.091$ & $\mathllap\shortminus0.150\pm0.206$ & $0.009\pm0.009$\\
        \bottomrule
    \end{tabular}

    \label{table:overlay}
\end{table}

\begin{figure*}[t]
    \centering
        \includegraphics[width=1\textwidth, height = 4.5cm, keepaspectratio]{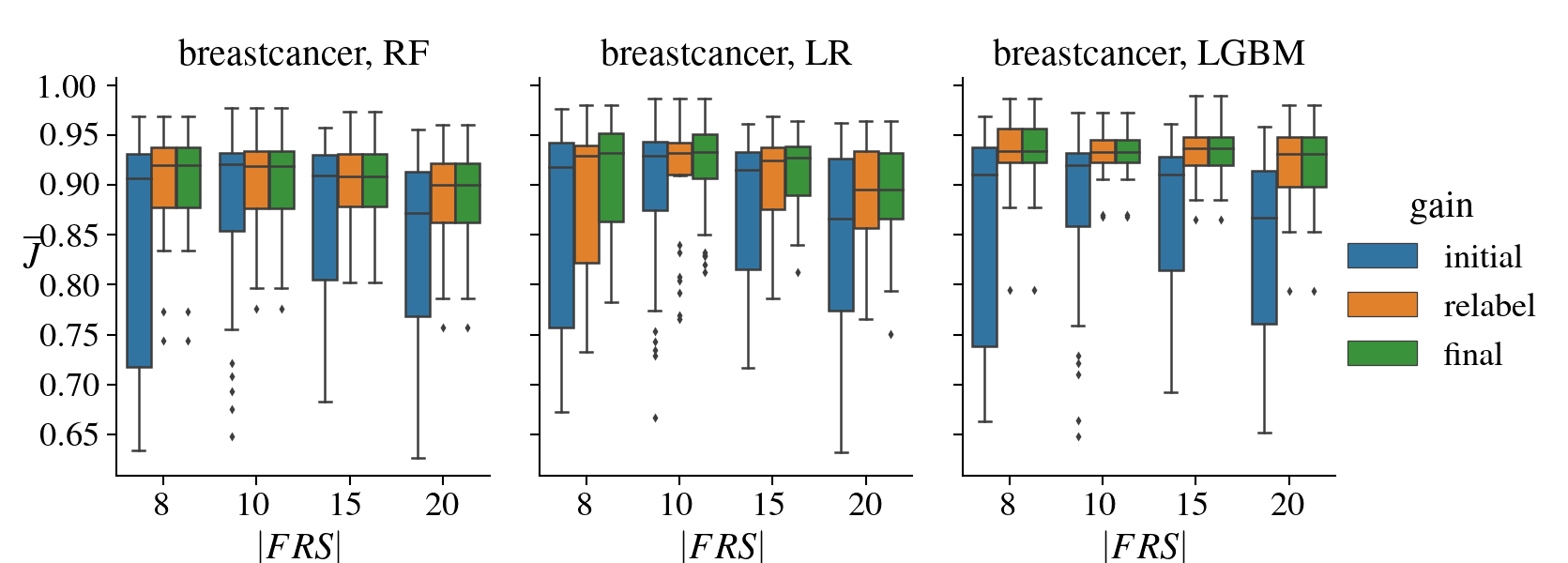}
     \caption{Effect of feedback rule set size for the Breast Cancer dataset and $\BPrand$ selection strategy. The same comparison as in Figure~\ref{fig:modperf} is shown between \textit{initial}, after \modrelabel, and \textit{final} (after FROTE). Each box and whiskers is computed from $30$ runs with $tcf=0.2$. }
    \label{fig:frs_len}
\end{figure*}

\textbf{Number of feedback rules.} One advantage of FROTE is its capability to work with rule sets containing any number of rules.  Figure~\ref{fig:frs_len} displays $\overline{J}$ values in the same manner as Figure~\ref{fig:modperf} for feedback rule sets having $8,10, 15$ and $20$ rules. The improvement in $\overline{J}$ is maintained up to $20$ rules. 
Results with other datasets are provided in Section B. Overall, they demonstrate the efficacy of FROTE with larger rule sets.

\begin{table}[ht]
    \small
    \centering
    \caption{Comparison of \BPrand{} and \BPIP{} base instance selection strategies. Means and standard deviations computed from all runs for a given dataset and model.}
    \vspace*{2mm}
    \begin{tabular}{@{}llcc@{}}
        \toprule
        Dataset & Model & \multicolumn{2}{c}{$\Delta \overline{J}$}\\ 
        \cmidrule(lr){3-4}
         & & \BPrand & \BPIP\\
        \midrule
        B.~Cancer &RF & $0.000\pm0.003 $ & $0.001\pm0.006 $\\
        &LR & $0.006\pm0.022 $ & $0.006\pm0.026 $\\
        &LGBM & $0.001\pm0.008 $ & $0.002\pm0.010 $\\\hline
        Car &RF & $0.005\pm0.020 $ & $0.006\pm0.020 $\\
        &LR & $0.022\pm0.034 $ & $0.020\pm0.029 $\\
        &LGBM & $0.008\pm0.033 $ & $0.008\pm0.027 $\\\hline
        Mushroom &RF & $0.001\pm0.017 $ & $0.004\pm0.034 $\\
        &LR & $0.005\pm0.023 $ & $0.011\pm0.049 $\\
        &LGBM & $0.004\pm0.037 $ & $0.006\pm0.041 $\\\hline
        Adult &RF & $0.003\pm0.014 $ & $0.003\pm0.011 $\\
        &LR & $0.008\pm0.023 $ & $0.004\pm0.012 $\\
        &LGBM & $0.004\pm0.015 $ & $0.003\pm0.011 $\\\hline
        Wine &RF & $0.001\pm0.007 $ & $0.001\pm0.007 $\\
        &LR & $0.056\pm0.096 $ & $0.055\pm0.094 $\\
        &LGBM & $0.003\pm0.015 $ & $0.003\pm0.010 $\\\hline
        Contracep. &RF & $0.032\pm0.081 $ & $0.038\pm0.085 $\\
        &LR & $0.041\pm0.099 $ & $0.051\pm0.102 $\\
        &LGBM & $0.027\pm0.066 $ & $0.026\pm0.057 $\\\hline
        Nursery &RF & $0.031\pm0.099 $ & $0.023\pm0.076 $\\
        &LR & $0.043\pm0.088 $ & $0.029\pm0.069 $\\
        &LGBM & $0.035\pm0.108 $ & $0.030\pm0.096 $\\\hline
        Splice &RF & $0.003\pm0.017 $ & $0.002\pm0.012 $\\
        &LR & $0.011\pm0.031 $ & $0.007\pm0.018 $\\
        &LGBM & $0.014\pm0.049 $ & $0.009\pm0.037 $\\
        \bottomrule
    \end{tabular}
    
    \label{table:IPvsAll}
\end{table}

\textbf{Base instance selection strategy.}
We now compare the performance of the two base instance selection strategies, $\BPrand$ and $\BPIP$. Table~\ref{table:IPvsAll} shows the $\overline{J}$ improvements for models trained on the final augmented dataset relative to the initial dataset. The amount of augmentation required (as a fraction of the input dataset size) for these improvements for both strategies is included in Section B. There is not a clear winner between $\BPrand$ and $\BPIP$ in terms of $\overline{J}$ (the ``win-loss-tie'' record based on 3 decimal places is 11-8-5), although $\BPIP$ generally adds fewer instances to the dataset. 
One possible reason behind relatively good performance of $\BPrand$ is although \BPIP{} appears more informed, \BPrand{} may avoid ``overfitting'', in the sense of selecting base instances that improve the objective function evaluated on the augmented training dataset but not on the held-out test set. 
Looking at the MRA and F-Score separately (provided in Section B) for the results in Table~\ref{table:IPvsAll}, we see an improvement in MRA without significant decrease (in some cases an increase) in F-Score for both techniques, for all results. However, the degree of improvement is dependent on the dataset and model.


\section{Broader Impact and Discussion}
One important point to note is that there is generally an inflection point in terms of the number of data points added where the cost to overall model performance starts to outweigh the improvement in MRA. This inflection point also depends on the model used and the dataset. It can be explained by the \textit{data difficulty factors} described in \cite{Stefanowski2016}, namely \textit{an effect of too strong overlap between classes}, and \textit{a presence of too many examples of one class inside the other class's region}. 

One limitation of the work is that it may be restricted to tabular data, however, we believe similar mechanisms could be used when considering images where Boolean rules could show relevant images segments or features. 


Our work supports model editing where the final ML model will encode the decision processes of not just the underlying data but also external knowledge. This ability can be leveraged to correct incorrect assumptions in the original data or encoded updated policies. On the other hand this introduces the ability for the model builder to influence the model outcomes which could intentionally or unintentionally introduce bias. The user feedback however is interpretable and transparent and user influence is in the form of a Boolean feedback rule. This supports easy integrating into a governance framework such as proposed in \cite{arnold2019factsheets} where clear auditing of the original data, the feedback rules and the newly created dataset can be stored to transparently log the updates to the model and capture the lineage of the data. Post processing analysis to compare the original and the resulting model could also be leveraged to ensure unintended biases have not been introduces \cite{bellamy2019ai} along with generating an interpretable model comparison of the two models as proposed by Nair et. al. \cite{nair2021ijcai}. Additionally, FROTE achieves this while trying to minimise the model accuracy for other segments of the dataset. This is in contrast to human labelling or relabelling tasks where the downstream impact of the newly labeled data points may be unclear. Additionally, the source of the newly labeled data, their level or expertise, familiarity with the data are all opaque. One could argue peer reviewing a feedback rule set to obtain consensus among stake holders is relatively easy compared to ensuring a consistent view is being used among data labellers.

\section{Conclusion}
We presented the problem of pre-processing training dataset to edit an ML model based on feedback rules. We proposed FROTE, a novel technique based on data augmentation, to solve this problem. Empirical studies on real datasets with different ML models demonstrate its effectiveness. Our work supports model editing where the final model encodes decision processes of not just the underlying data but also external knowledge. This ability can be leveraged to correct deficiencies in the original data or adapt to updated policies. User feedback is interpretable and transparent as it is in the form of Boolean rules, supporting clear auditing and governance. A promising future direction is to experiment on different base population selection strategies and optimization techniques to select the base instances and their neighbors together, in order to improve the performance.  

\bibliography{references}

\begin{thebibliography}{34}
\providecommand{\natexlab}[1]{#1}
\providecommand{\url}[1]{\texttt{#1}}
\expandafter\ifx\csname urlstyle\endcsname\relax
  \providecommand{\doi}[1]{doi: #1}\else
  \providecommand{\doi}{doi: \begingroup \urlstyle{rm}\Url}\fi

\bibitem[Arnold et~al.(2019)Arnold, Bellamy, Hind, Houde, Mehta,
  Mojsilovi{\'c}, Nair, Ramamurthy, Olteanu, Piorkowski,
  et~al.]{arnold2019factsheets}
Arnold, M., Bellamy, R.~K., Hind, M., Houde, S., Mehta, S., Mojsilovi{\'c}, A.,
  Nair, R., Ramamurthy, K.~N., Olteanu, A., Piorkowski, D., et~al.
\newblock Factsheets: Increasing trust in ai services through supplier's
  declarations of conformity.
\newblock \emph{IBM Journal of Research and Development}, 63\penalty0
  (4/5):\penalty0 6--1, 2019.

\bibitem[Awasthi et~al.(2020)Awasthi, Ghosh, Goyal, and
  Sarawagi]{awasthi2020learning}
Awasthi, A., Ghosh, S., Goyal, R., and Sarawagi, S.
\newblock Learning from rules generalizing labeled exemplars.
\newblock \emph{arXiv preprint arXiv:2004.06025}, 2020.

\bibitem[Bellamy et~al.(2019)Bellamy, Dey, Hind, Hoffman, Houde, Kannan, Lohia,
  Martino, Mehta, Mojsilovi{\'c}, et~al.]{bellamy2019ai}
Bellamy, R.~K., Dey, K., Hind, M., Hoffman, S.~C., Houde, S., Kannan, K.,
  Lohia, P., Martino, J., Mehta, S., Mojsilovi{\'c}, A., et~al.
\newblock Ai fairness 360: An extensible toolkit for detecting and mitigating
  algorithmic bias.
\newblock \emph{IBM Journal of Research and Development}, 63\penalty0
  (4/5):\penalty0 4--1, 2019.

\bibitem[Cakmak et~al.(2010)Cakmak, Chao, and Thomaz]{cakmak2010designing}
Cakmak, M., Chao, C., and Thomaz, A.~L.
\newblock Designing interactions for robot active learners.
\newblock \emph{IEEE Transactions on Autonomous Mental Development}, 2\penalty0
  (2):\penalty0 108--118, 2010.

\bibitem[Calders et~al.(2009)Calders, Kamiran, and
  Pechenizkiy]{calders2009building}
Calders, T., Kamiran, F., and Pechenizkiy, M.
\newblock Building classifiers with independency constraints.
\newblock In \emph{2009 IEEE International Conference on Data Mining
  Workshops}, pp.\  13--18. IEEE, 2009.

\bibitem[Calmon et~al.(2017)Calmon, Wei, Vinzamuri, Ramamurthy, and
  Varshney]{optimizedpre}
Calmon, F.~P., Wei, D., Vinzamuri, B., Ramamurthy, K.~N., and Varshney, K.~R.
\newblock Optimized pre-processing for discrimination prevention.
\newblock In \emph{Proceedings of the 31st International Conference on Neural
  Information Processing Systems}, NIPS'17, pp.\  3995–4004, Red Hook, NY,
  USA, 2017. Curran Associates Inc.
\newblock ISBN 9781510860964.

\bibitem[Chawla et~al.(2002)Chawla, Bowyer, Hall, and Kegelmeyer]{SMOTE_orig}
Chawla, N.~V., Bowyer, K.~W., Hall, L.~O., and Kegelmeyer, W.~P.
\newblock Smote: Synthetic minority over-sampling technique.
\newblock \emph{J. Artif. Int. Res.}, 16\penalty0 (1):\penalty0 321–357, June
  2002.
\newblock ISSN 1076-9757.

\bibitem[Dai et~al.(2007)Dai, Yang, Xue, and Yu]{dai2007boosting}
Dai, W., Yang, Q., Xue, G.-R., and Yu, Y.
\newblock Boosting for transfer learning.
\newblock In \emph{Proceedings of the 24th international conference on Machine
  learning}, pp.\  193--200, 2007.

\bibitem[Daly et~al.(2021)Daly, Mattetti, Alkan, and Nair]{daly2021aaai}
Daly, E.~M., Mattetti, M., Alkan, {\"O}., and Nair, R.
\newblock User driven model adjustment via boolean rule explanations.
\newblock In \emph{AAAI 2021}, 2021.

\bibitem[Dash et~al.(2018)Dash, G\"{u}nl\"{u}k, and Wei]{brcg2018}
Dash, S., G\"{u}nl\"{u}k, O., and Wei, D.
\newblock Boolean decision rules via column generation.
\newblock In \emph{Proceedings of the 32nd International Conference on Neural
  Information Processing Systems}, NIPS’18, pp.\  4660–4670, Red Hook, NY,
  USA, 2018. Curran Associates Inc.

\bibitem[Douzas \& Bacao(2018)Douzas and Bacao]{douzas2018effective}
Douzas, G. and Bacao, F.
\newblock Effective data generation for imbalanced learning using conditional
  generative adversarial networks.
\newblock \emph{Expert Systems with applications}, 91:\penalty0 464--471, 2018.

\bibitem[Eaton \& desJardins(2011)Eaton and desJardins]{eaton2011selective}
Eaton, E. and desJardins, M.
\newblock Selective transfer between learning tasks using task-based boosting.
\newblock In \emph{AAAI}, 2011.

\bibitem[Fern\'{a}ndez et~al.(2018)Fern\'{a}ndez, Garc\'{\i}a, Herrera, and
  Chawla]{smoteAnniversary}
Fern\'{a}ndez, A., Garc\'{\i}a, S., Herrera, F., and Chawla, N.~V.
\newblock Smote for learning from imbalanced data: Progress and challenges,
  marking the 15-year anniversary.
\newblock \emph{J. Artif. Int. Res.}, 61\penalty0 (1):\penalty0 863–905,
  January 2018.
\newblock ISSN 1076-9757.

\bibitem[Goodfellow et~al.(2014)Goodfellow, Pouget-Abadie, Mirza, Xu,
  Warde-Farley, Ozair, Courville, and Bengio]{goodfellow2014generative}
Goodfellow, I., Pouget-Abadie, J., Mirza, M., Xu, B., Warde-Farley, D., Ozair,
  S., Courville, A., and Bengio, Y.
\newblock Generative adversarial nets.
\newblock In \emph{Advances in Neural Information Processing Systems}, pp.\
  2672--2680, 2014.

\bibitem[Guillory \& Bilmes(2011)Guillory and Bilmes]{guillory2011simultaneous}
Guillory, A. and Bilmes, J.~A.
\newblock Simultaneous learning and covering with adversarial noise.
\newblock In \emph{ICML}, 2011.

\bibitem[Han et~al.(2005)Han, Wang, and Mao]{han2005borderline}
Han, H., Wang, W.-Y., and Mao, B.-H.
\newblock Borderline-smote: a new over-sampling method in imbalanced data sets
  learning.
\newblock In \emph{International Conference on Intelligent Computing}, pp.\
  878--887. Springer, 2005.

\bibitem[Kapoor et~al.(2010)Kapoor, Lee, Tan, and
  Horvitz]{kapoor2010interactive}
Kapoor, A., Lee, B., Tan, D., and Horvitz, E.
\newblock Interactive optimization for steering machine classification.
\newblock In \emph{Proceedings of the SIGCHI Conference on Human Factors in
  Computing Systems}, pp.\  1343--1352, 2010.

\bibitem[Ke et~al.(2017)Ke, Meng, Finley, Wang, Chen, Ma, Ye, and
  Liu]{ke2017lightgbm}
Ke, G., Meng, Q., Finley, T., Wang, T., Chen, W., Ma, W., Ye, Q., and Liu,
  T.-Y.
\newblock Lightgbm: A highly efficient gradient boosting decision tree.
\newblock \emph{Advances in neural information processing systems},
  30:\penalty0 3146--3154, 2017.

\bibitem[Khandani et~al.(2010)Khandani, Kim, and Lo]{khandani2010consumer}
Khandani, A.~E., Kim, A.~J., and Lo, A.~W.
\newblock Consumer credit-risk models via machine-learning algorithms.
\newblock \emph{Journal of Banking \& Finance}, 34\penalty0 (11):\penalty0
  2767--2787, 2010.

\bibitem[Lakkaraju et~al.(2016)Lakkaraju, Bach, and
  Leskovec]{Interpretable_Decision_Sets_KDD2016}
Lakkaraju, H., Bach, S.~H., and Leskovec, J.
\newblock Interpretable decision sets: A joint framework for description and
  prediction.
\newblock In \emph{Proceedings of the 22nd ACM SIGKDD International Conference
  on Knowledge Discovery and Data Mining}, KDD ’16, pp.\  1675–1684, New
  York, NY, USA, 2016. Association for Computing Machinery.
\newblock ISBN 9781450342322.

\bibitem[Lauer \& Bloch(2008)Lauer and Bloch]{LAUER20081578}
Lauer, F. and Bloch, G.
\newblock Incorporating prior knowledge in support vector machines for
  classification: A review.
\newblock \emph{Neurocomputing}, 71\penalty0 (7):\penalty0 1578--1594, 2008.
\newblock ISSN 0925-2312.
\newblock \doi{https://doi.org/10.1016/j.neucom.2007.04.010}.
\newblock URL
  \url{https://www.sciencedirect.com/science/article/pii/S0925231207001439}.
\newblock Progress in Modeling, Theory, and Application of Computational
  Intelligenc.

\bibitem[Letham et~al.(2015)Letham, Rudin, McCormick, and
  Madigan]{interpretablerules2015}
Letham, B., Rudin, C., McCormick, T., and Madigan, D.
\newblock Interpretable classifiers using rules and bayesian analysis: Building
  a better stroke prediction model.
\newblock \emph{The Annals of Applied Statistics}, 9:\penalty0 1350--1371, 09
  2015.
\newblock \doi{10.1214/15-AOAS848}.

\bibitem[Maclin et~al.(2006)Maclin, Shavlik, Walker, and
  Torrey]{SVM_inc_rules_2006}
Maclin, R., Shavlik, J., Walker, T., and Torrey, L.
\newblock A simple and effective method for incorporating advice into kernel
  methods.
\newblock In \emph{Proceedings, The Twenty-First National Conference on
  Artificial Intelligence and the Eighteenth Innovative Applications of
  Artificial Intelligence Conference}, 01 2006.

\bibitem[Molnar(2019)]{molnar2019}
Molnar, C.
\newblock \emph{Interpretable Machine Learning}.
\newblock 2019.
\newblock \url{https://christophm.github.io/interpretable-ml-book/}.

\bibitem[Nair et~al.(2021)Nair, Mattetti, Daly, Wei, Alkan, and
  Zhang]{nair2021ijcai}
Nair, R., Mattetti, M., Daly, E., Wei, D., Alkan, O., and Zhang, Y.
\newblock What changed? interpretable model comparison.
\newblock In \emph{IJCAI 2021}, 2021.

\bibitem[Pedregosa et~al.(2011)Pedregosa, Varoquaux, Gramfort, Michel, Thirion,
  Grisel, Blondel, Prettenhofer, Weiss, Dubourg, Vanderplas, Passos,
  Cournapeau, Brucher, Perrot, and Duchesnay]{scikit-learn}
Pedregosa, F., Varoquaux, G., Gramfort, A., Michel, V., Thirion, B., Grisel,
  O., Blondel, M., Prettenhofer, P., Weiss, R., Dubourg, V., Vanderplas, J.,
  Passos, A., Cournapeau, D., Brucher, M., Perrot, M., and Duchesnay, E.
\newblock Scikit-learn: Machine learning in {P}ython.
\newblock \emph{Journal of Machine Learning Research}, 12:\penalty0 2825--2830,
  2011.

\bibitem[Ratner et~al.(2017)Ratner, Bach, Ehrenberg, Fries, Wu, and
  R{\'e}]{ratner2017snorkel}
Ratner, A., Bach, S.~H., Ehrenberg, H., Fries, J., Wu, S., and R{\'e}, C.
\newblock Snorkel: Rapid training data creation with weak supervision.
\newblock In \emph{Proceedings of the VLDB Endowment. International Conference
  on Very Large Data Bases}, volume~11, pp.\  269. NIH Public Access, 2017.

\bibitem[Ribeiro et~al.(2018)Ribeiro, Singh, and Guestrin]{ribeiro2018anchors}
Ribeiro, M.~T., Singh, S., and Guestrin, C.
\newblock Anchors: High-precision model-agnostic explanations.
\newblock In \emph{Proceedings of the AAAI Conference on Artificial
  Intelligence}, volume~32, 2018.

\bibitem[Sharma et~al.(2020)Sharma, Zhang, R\'{\i}os~Aliaga, Bouneffouf,
  Muthusamy, and Varshney]{AIES_DataAug_Fairness_2020}
Sharma, S., Zhang, Y., R\'{\i}os~Aliaga, J.~M., Bouneffouf, D., Muthusamy, V.,
  and Varshney, K.~R.
\newblock Data augmentation for discrimination prevention and bias
  disambiguation.
\newblock In \emph{Proceedings of the AAAI/ACM Conference on AI, Ethics, and
  Society}, AIES ’20, pp.\  358–364, New York, NY, USA, 2020. Association
  for Computing Machinery.
\newblock ISBN 9781450371100.
\newblock \doi{10.1145/3375627.3375865}.
\newblock URL \url{https://doi.org/10.1145/3375627.3375865}.

\bibitem[Singh \& Urolagin(2020)Singh and Urolagin]{singhuse}
Singh, J. and Urolagin, S.
\newblock Use of artificial intelligence for health insurance claims
  automation.
\newblock In \emph{Advances in Machine Learning and Computational
  Intelligence}, pp.\  381--392. Springer, 2020.

\bibitem[Stefanowski(2016)]{Stefanowski2016}
Stefanowski, J.
\newblock \emph{Dealing with Data Difficulty Factors While Learning from
  Imbalanced Data}, pp.\  333--363.
\newblock Springer International Publishing, Cham, 2016.
\newblock ISBN 978-3-319-18781-5.
\newblock \doi{10.1007/978-3-319-18781-5_17}.
\newblock URL \url{https://doi.org/10.1007/978-3-319-18781-5_17}.

\bibitem[Tanaka \& Aranha(2019)Tanaka and Aranha]{tanaka2019data}
Tanaka, F. H. K. d.~S. and Aranha, C.
\newblock Data augmentation using gans.
\newblock \emph{arXiv preprint arXiv:1904.09135}, 2019.

\bibitem[Xu et~al.(2019)Xu, Skoularidou, Cuesta-Infante, and
  Veeramachaneni]{CTGAN}
Xu, L., Skoularidou, M., Cuesta-Infante, A., and Veeramachaneni, K.
\newblock Modeling tabular data using conditional gan.
\newblock In \emph{Advances in Neural Information Processing Systems}, 2019.

\bibitem[Zhang \& Deng(2015)Zhang and Deng]{interprrules2017}
Zhang, Y. and Deng, A.
\newblock Redundancy rules reduction in rule-based knowledge bases.
\newblock pp.\  639--643, 08 2015.
\newblock \doi{10.1109/FSKD.2015.7382017}.

\end{thebibliography}
\bibliographystyle{mlsys2022}

\appendix
\section{Solution Details}

\paragraph{Subset selection via integer programming.}
We elaborate on the integer programming formulation for the subset selection problem presented in the main paper. For a given $\FRS$ we would like to determine a subset of the training data within $\FRS$ that has the greatest influence on the model decision boundaries. 

The weight $w_i$ reflects the value of a data point for the final selection. Instances near the decision boundary are more valuable, as it has a greater potentially to influence the model. This weight is pre-computed as follows:

For each $j,\, j\in D$ compute, $p$ as the number of $k$ neighbours who have the same label, and $q$ as the number of $k$ neighbours who have a different label. Here, the label refers to the predicted label from a model we seek to edit. If $q>>p$, the observation can be considered \emph{noisy}, $p>>q$, then the observation can be considered as \emph{safe}, and if $p\approx q$ the observation can be considered as \emph{borderline}~\cite{han2005borderline}. Correspondingly, the weights $w_i$ can be assigned based on these three cases such that, the borderline points is assigned the largest weight. In our experiments, we set $w_i=3$ for borderline and $w_i = 1$ for noisy and safe data points computed within $k=10$ nearest neighbors.

\paragraph{Subset selection with online learning.}

As mentioned in the main text, we also considered the use of online learning to simplify the evaluation of objective function~(3), and specifically to avoid running training algorithm $A$. We instead take a proxy approach in which 1) the current model $M_{\hat{D}} = A(\hat{D})$ is approximated by a (parametric) model $\hat{M}$ to which online learning can be applied, and 2) the retrained model $M_{D'}$ is approximated by the result of online learning, starting from $\hat{M}$ and updating based on the generated instances $\mathrm{Generate}(\mathcal{B})$. Recalling that $J$ is also replaced by its empirical approximation $\hat{J}_{\hat{D}}$ over $\hat{D}$, the online learning approximation can thus be written as 

\begin{equation}
\begin{aligned}\label{eqn:OLproxy}
&J\left(A\left(\hat{D} \cup \mathrm{Generate}(\mathcal{B})\right), \mathcal{F}\right) \approx\\ &\hat{J}_{\hat{D}}\left( \mathrm{OL}\left(\hat{M}, \mathrm{Generate}(\mathcal{B})\right), \mathcal{F} \right).
\end{aligned}
\end{equation}

We investigated the use of \ref{eqn:OLproxy} to approximate objective function (3) for singleton sets $\mathcal{B} = \{i\}$, $i \in \mathcal{P}$. Such evaluations on singletons might be summed to provide a crude approximation to (3) for non-singleton $\mathcal{B}$; the \textit{IP} objective function (4) is also a sum approximation in this sense. They could also constitute the first iteration in a greedy algorithm for selecting $\mathcal{B}$. 

Our experience thus far however is that even the evaluation of \eqref{eqn:OLproxy} is still too computationally intensive to be practical (at least in terms of facilitating experimentation). To be more specific, we used the Vowpal Wabbit library\footnote{\url{https://vowpalwabbit.org}} for online learning with a plain logistic regression model $\hat{M}$. Step 1) of approximating $M_{\hat{D}}$ with $\hat{M}$ is done by training $\hat{M}$ on dataset $\hat{D}$ and the outputs of $M_{\hat{D}}$ on $\hat{D}$. This has computational complexity $O(\lvert\hat{D}\rvert)$. Likewise, step 2), i.e.~approximating $M_{D'}$ by updating $\hat{M}$ for each generated instance $\mathrm{Generate}(\{i\})$, $i \in \mathcal{P}$, also has complexity $O(\lvert\mathcal{P}\rvert) = O(\lvert\hat{D}\rvert)$. However, evaluating $\hat{J}_{\hat{D}}$ for each of these updated models results in complexity $O(\lvert\hat{D}\rvert^2)$, and we have found this to be the slow step in our limited experiments. Future work could consider further approximations to the objective function that avoid higher than first-order complexity in $\lvert\hat{D}\rvert$.

\paragraph{Synthetic instance generation.}
Synthetic instance generation is used by the \textit{Generate()} procedure within FROTE, as outlined in Algorithm~1 of main paper (line 9). It is called for each \textit{base instance} and a randomly selected neighbor of it in order to generate synthetic instances. Synthetic instance generation uses two subroutines for populating categorical and numerical attributes. 

For populating categorical attributes, algorithm iterates through each categorical attribute to assign a value. For each categorical attribute, initially, all possible values for that attribute are calculated and stored. These attribute values are sorted in the decreasing order of the number of times they occur in the neighbors. Therefore, the first element in the list is the value that occurs in the majority of the $k$ nearest neighbor instances of the base instance. If the corresponding attribute is part of one of the conditions of the rule, then a special check is needed to make sure that the assigned value satisfies the corresponding condition(s). For instance, the algorithm ensures that for a condition with "$\ne$" operator, the value assigned to the corresponding attribute is different than the \textit{value} of that corresponding condition. 

The procedure iterates over each of the numerical attributes, and for each attribute, if it is not part of any of the conditions of the rule, the value to the corresponding numerical attribute is assigned using a similar approach to SMOTE~\cite{SMOTE_orig}. If the attribute exists in a condition where the operator is '$=$', then the value of the corresponding condition is assigned. 
However, if the attribute exists in a condition where the operator is one of \{'$>$','$\ge$','$<$','$\le$'\}, extra checks are performed to ensure that the generated value satisfies the corresponding conditions. Specifically, a window is defined with a minimum and maximum value (lines 21-29) based on the specific operators. These bounds keep track of the minimum and/or the maximum  values that can be assigned to the corresponding feature of the new instance. They are further adjusted based on the base and neighbor instance values  to make sure that the new value that will be assigned will stay within the value limits defined by the comparison operators. Finally a \textit{diff} value is assigned based on a tightest window determined by these minimum and maximum values together with the base and the neighbor instances' corresponding attribute values, and \textit{diff} is then used to generate a value for the corresponding attribute.

\section{Experimental Evaluation}

\subsection{Further Experimental Results}
\begin{figure*} 
    \centering
        \includegraphics[width=1\textwidth]{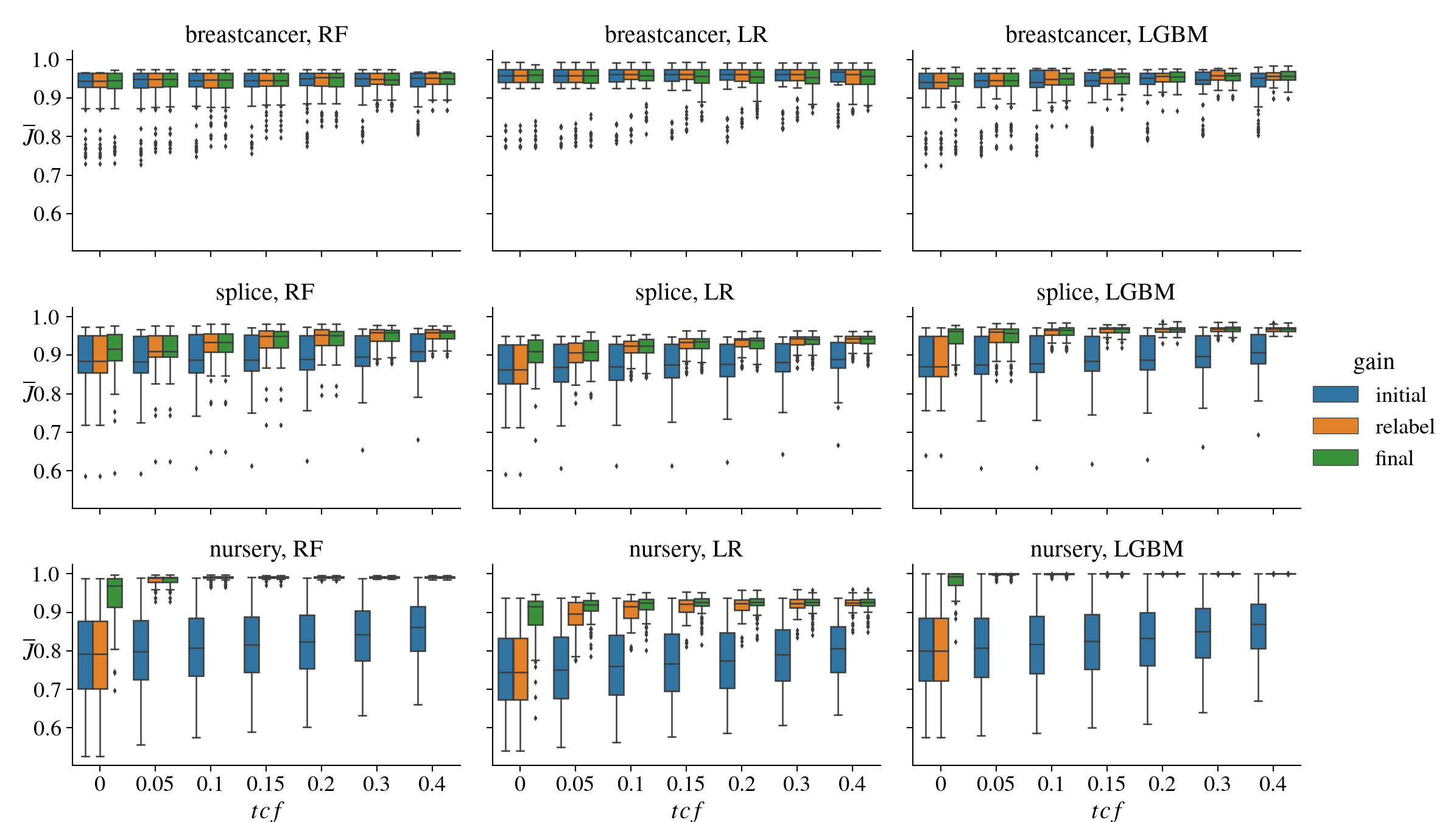}
        \includegraphics[width=1\textwidth]{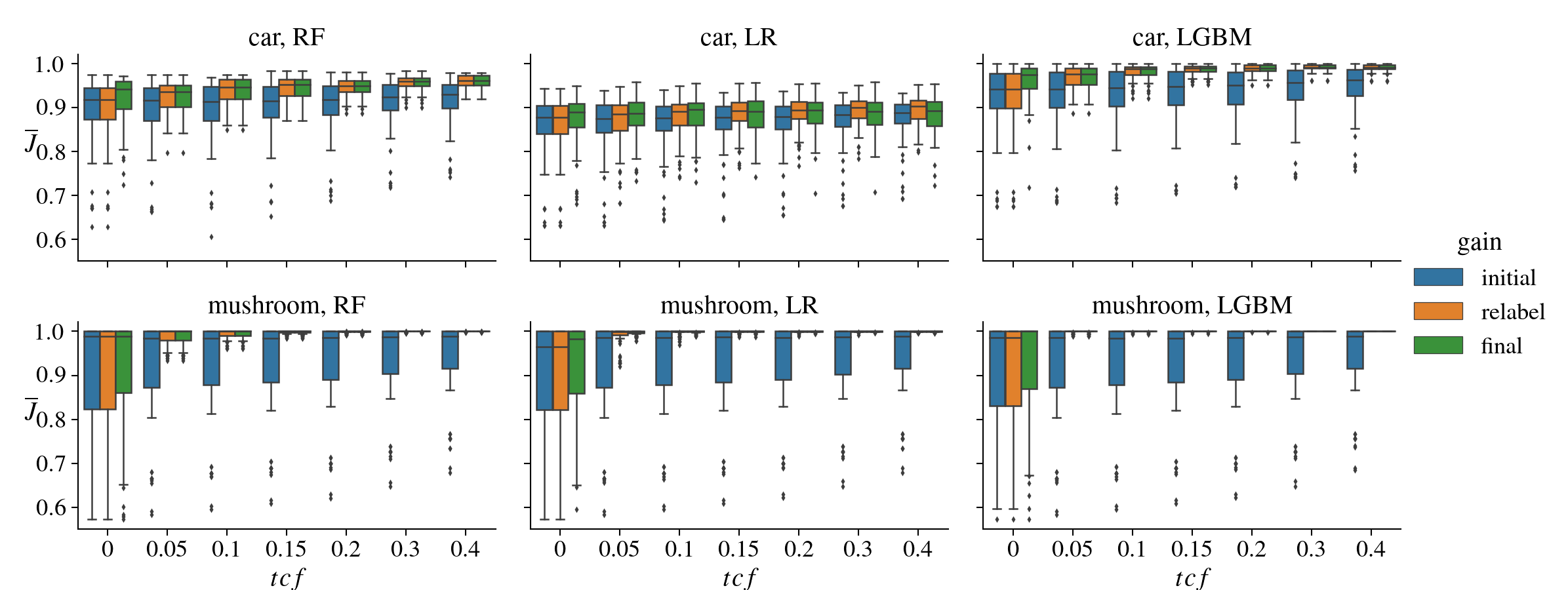}

     \caption{Additional plots for Figure~2 in the main paper. Experiments with models trained on the initial dataset before FROTE (\textit{initial}), after applying the \modrelabel{} mod strategy, and after FROTE completes augmentation (\textit{final}). The comparison is shown as a function of the training coverage fraction of the feedback rule sets and for different ML models and the Splice, Nursery and Breast Cancer, Mushroom and Car datasets. The $\BPrand$ selection strategy is used. Standard box plot showing interquartile range (IQR) and whiskers showing $1.5$ times IQR  based on 30 random draws for each of $\lvert\FRS\rvert \in \{1,3,5\}$.}
    \label{fig:fig_relabel}
\end{figure*}

\begin{figure*} [th]
    \centering
        \includegraphics[width=1\textwidth]{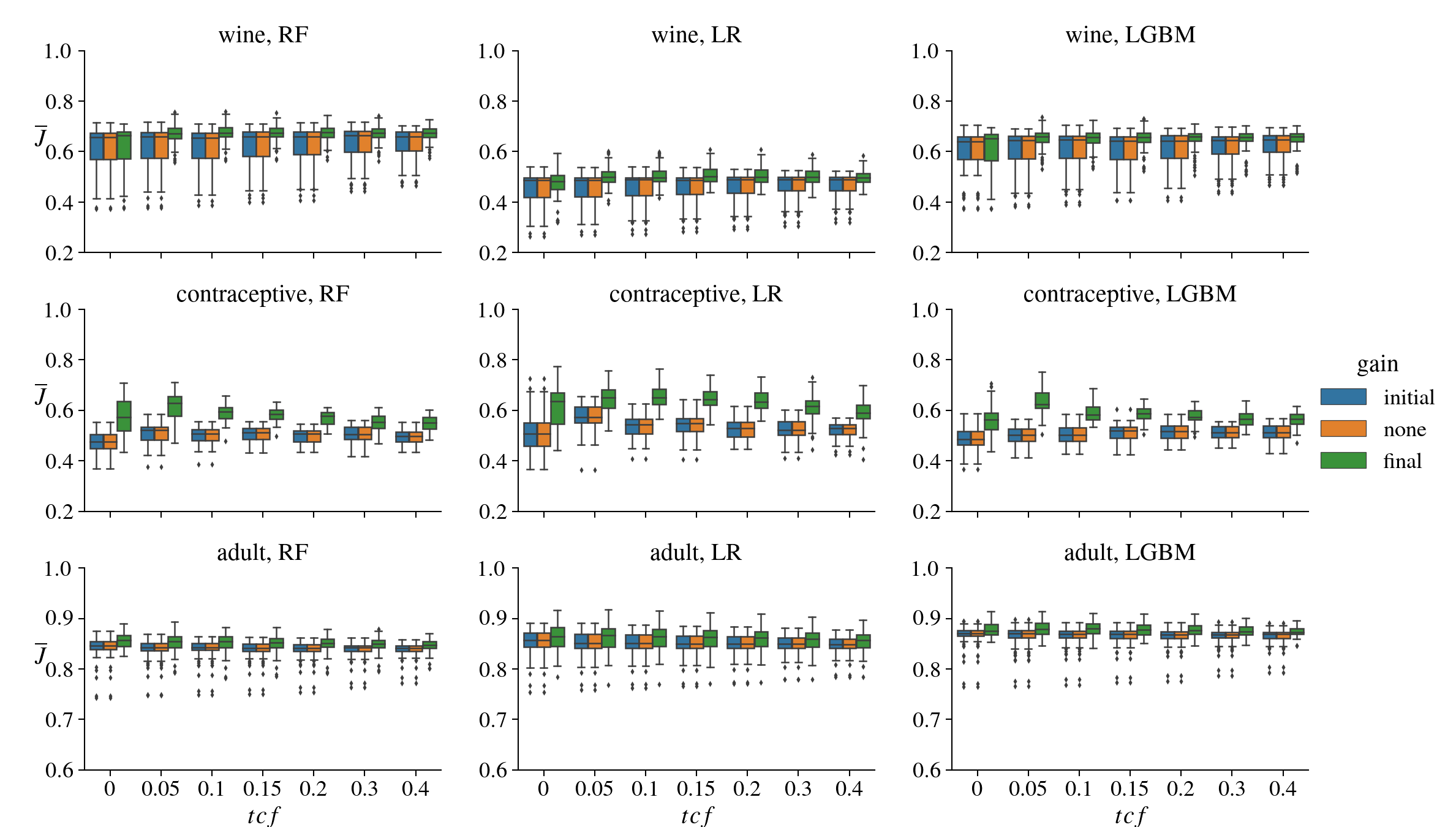}
     \caption{Additional plots for Figure~2 in the main paper. Experiments with models trained on the initial dataset before FROTE (\textit{initial}), after applying the \modnone{} strategy, and after FROTE completes augmentation (\textit{final}). $mod-imp$ and $final-imp$ represent the differences in $\overline{J}$ between \textit{mod} and \textit{initial} and \textit{final} and \textit{mod}, respectively. The comparison is shown as a function of the training coverage fraction of the feedback rule sets and for different ML models and all the datasets. The $\BPrand$ selection strategy is used. Standard box plot showing interquartile range (IQR) and whiskers showing $1.5$ times IQR  based on 30 random draws for each of $\lvert\FRS\rvert \in \{1,3,5\}$.}
    \label{fig:fignone1}
\end{figure*}

\begin{figure*} [th]
    \centering
        \includegraphics[width=1\textwidth]{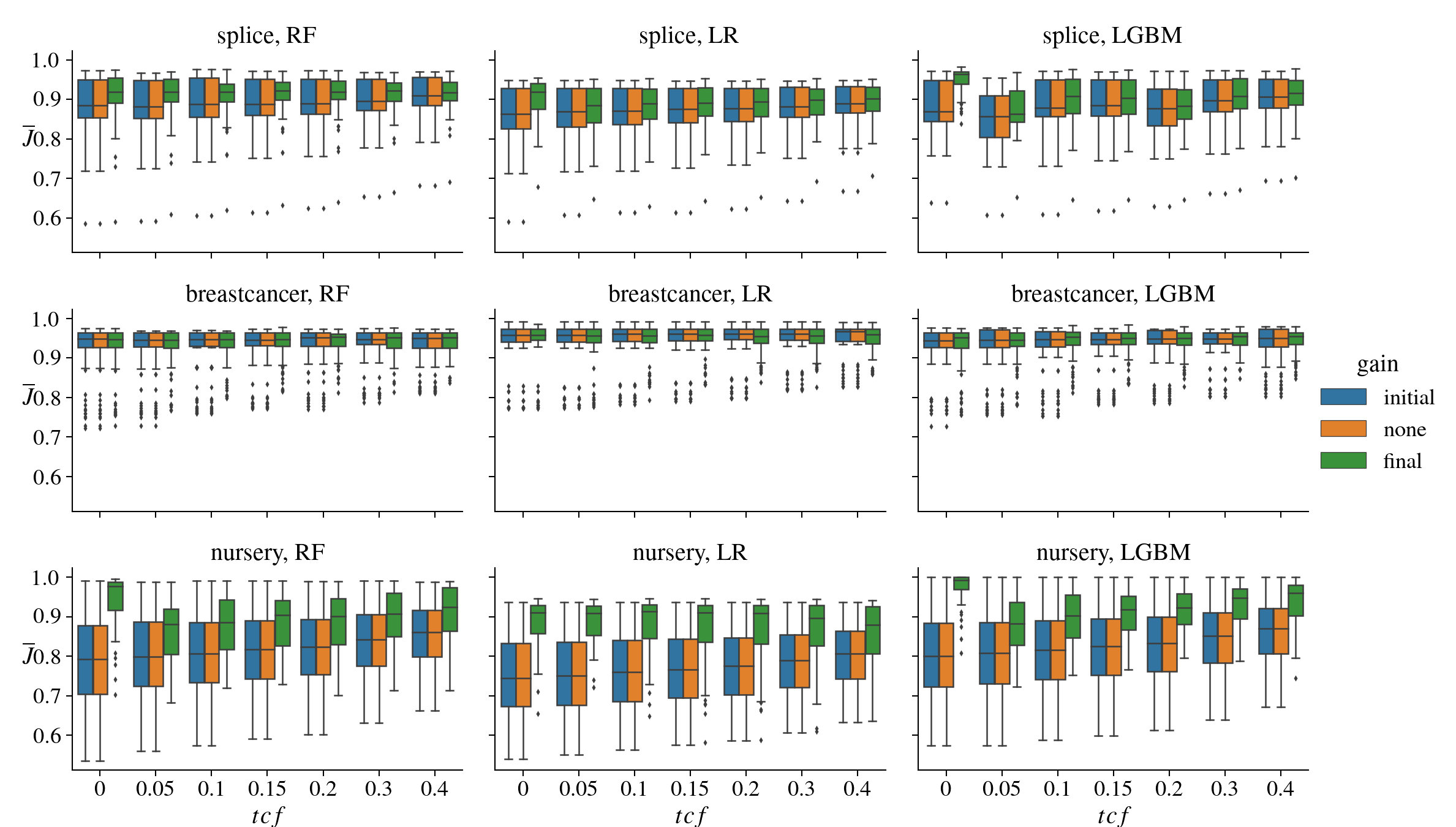}
        \includegraphics[width=1\textwidth]{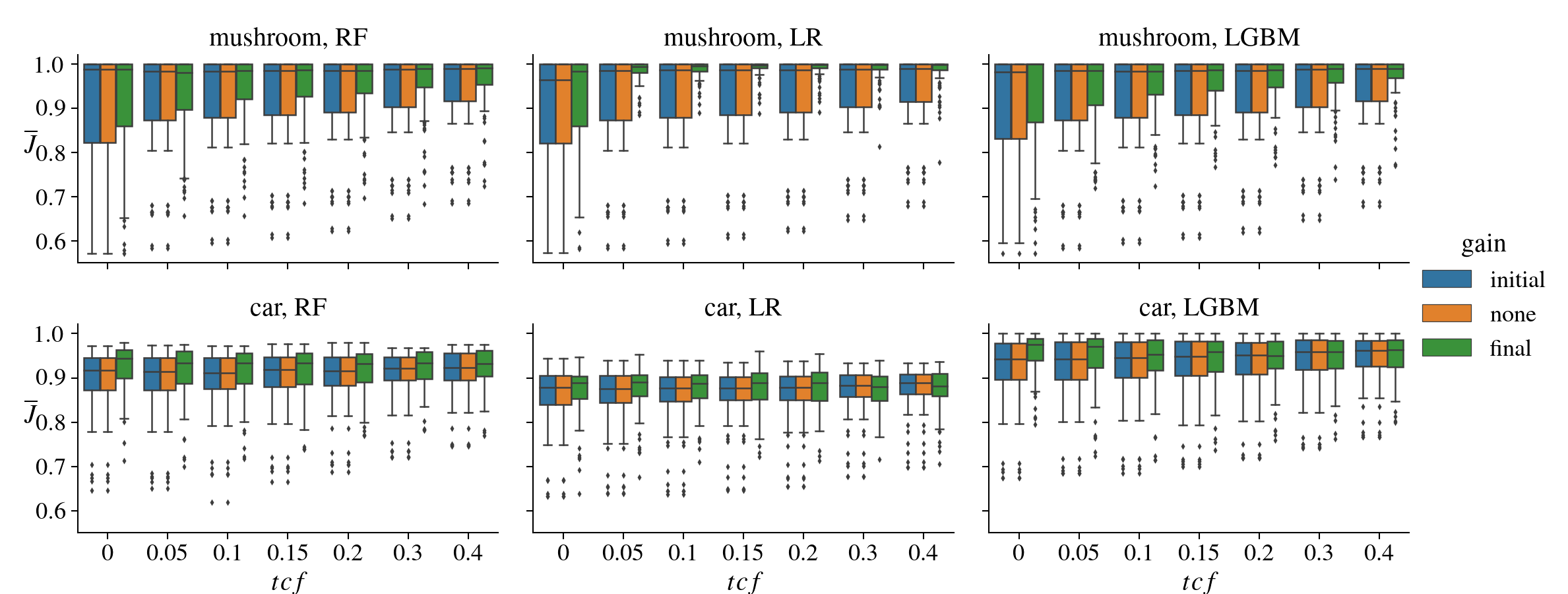}
     \caption{Additional plots for Figure~2 in the main paper. Experiments with models trained on the initial dataset before FROTE (\textit{initial}), after applying the \modnone{} strategy, and after FROTE completes augmentation (\textit{final}). $mod-imp$ and $final-imp$ represent the differences in $\overline{J}$ between \textit{mod} and \textit{initial} and \textit{final} and \textit{mod}, respectively. The comparison is shown as a function of the training coverage fraction of the feedback rule sets and for different ML models and all the datasets. The $\BPrand$ selection strategy is used. Standard box plot showing interquartile range (IQR) and whiskers showing $1.5$ times IQR  based on 30 random draws for each of $\lvert\FRS\rvert \in \{1,3,5\}$.}
    \label{fig:fignone2}
\end{figure*}

\begin{figure*} [th]
    \centering
        \includegraphics[width=1\textwidth]{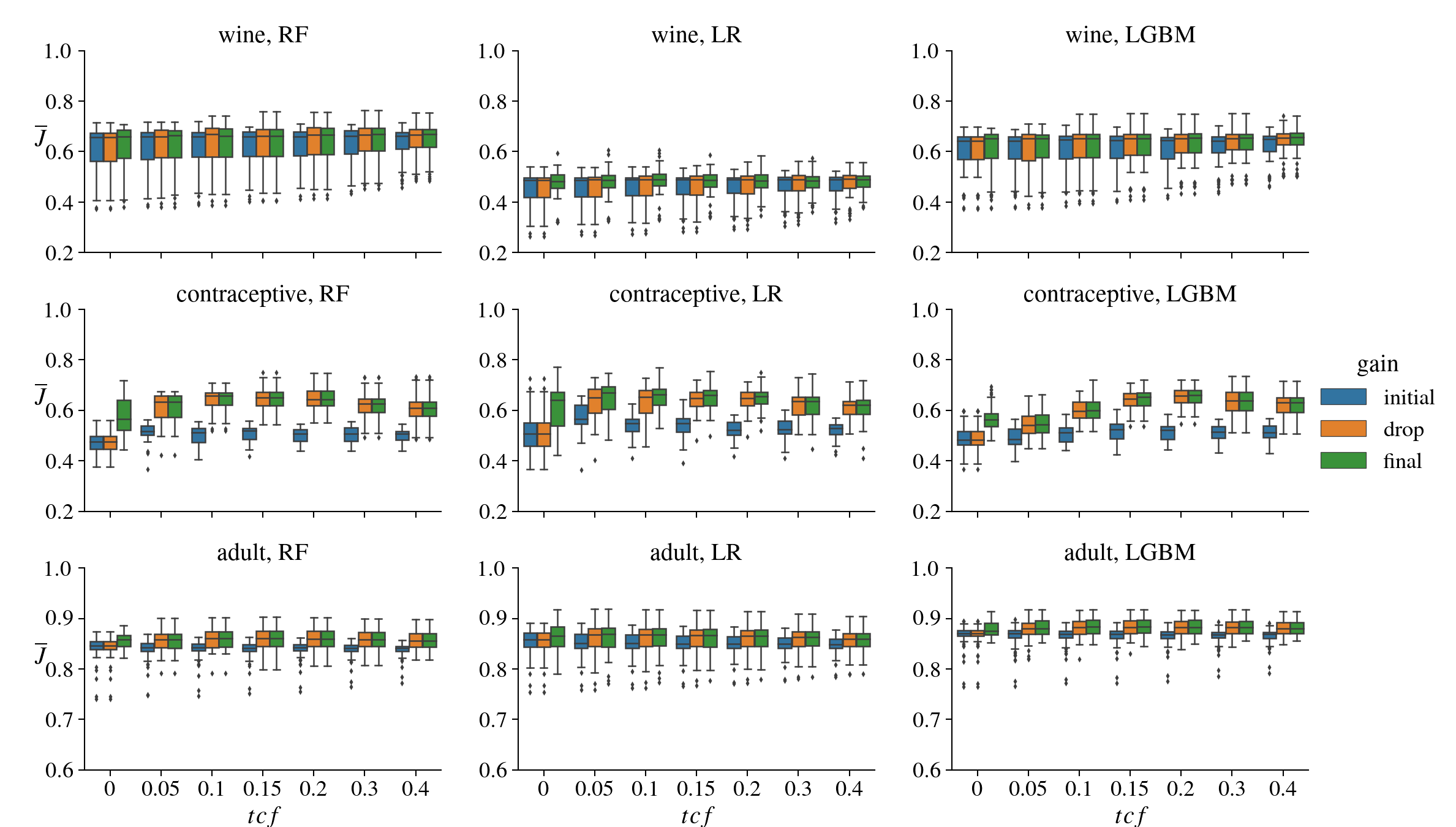}
     \caption{Similar setting with Figure 1 except results are presented for \moddrop{} modification strategy.}
    \label{fig:figdrop1}
\end{figure*}

\begin{figure*} [th]
    \centering
        \includegraphics[width=1\textwidth]{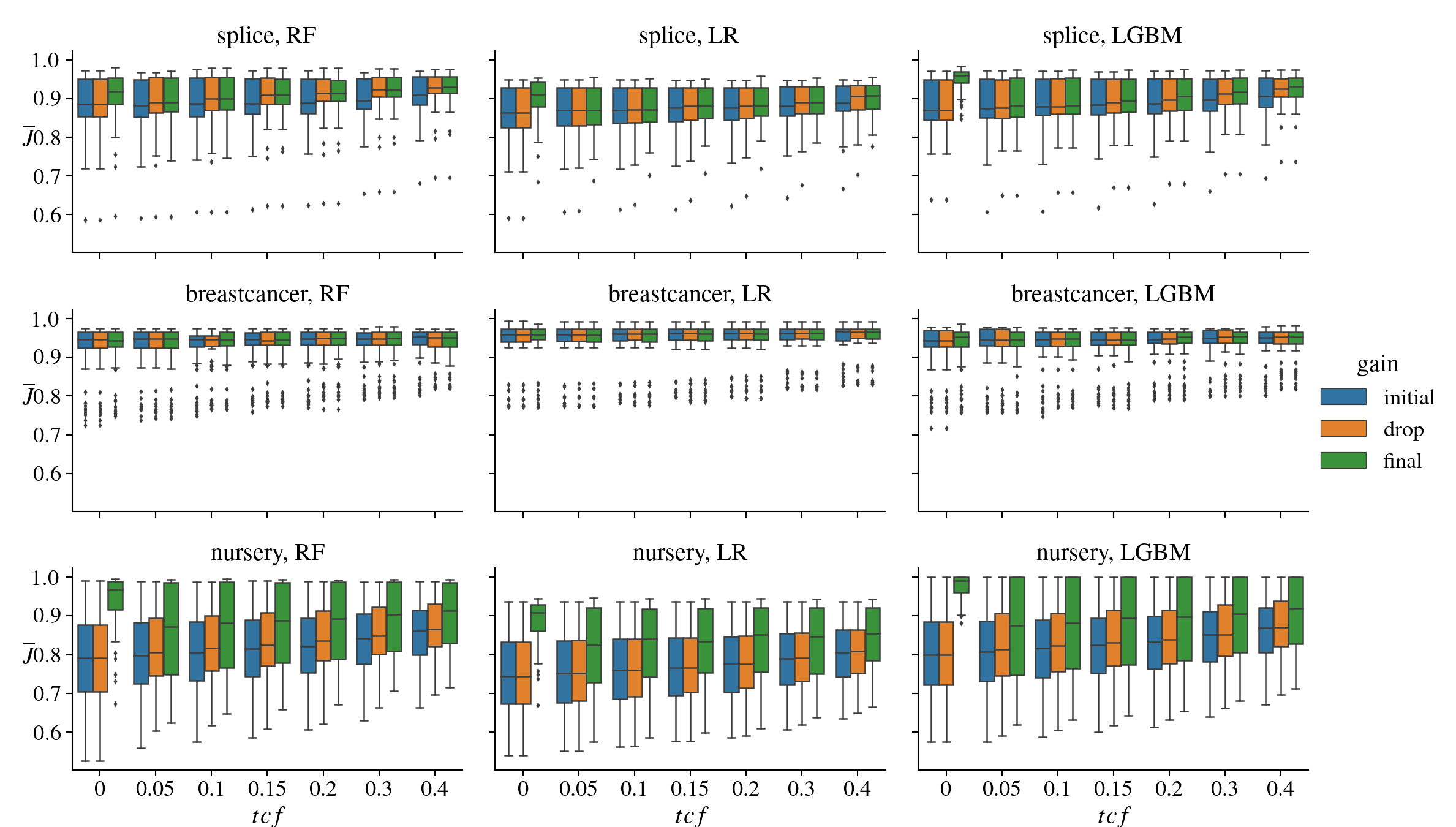}
        \includegraphics[width=1\textwidth]{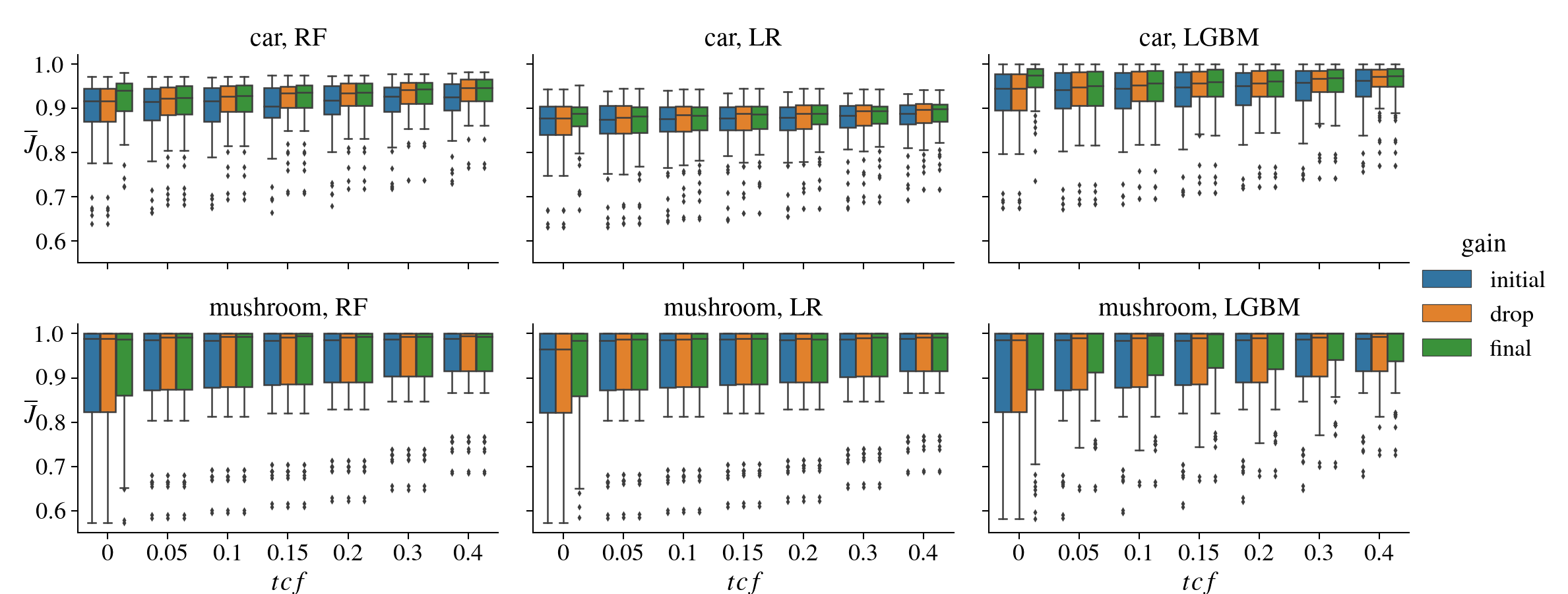}
     \caption{Similar setting with Figure 1 except results are presented for \moddrop{} modification strategy.}
    \label{fig:figdrop2}
\end{figure*}

\textbf{Benefit of augmentation.} We compare the test set $\overline{J}$ values obtained from the models that are
trained on 1) the initial dataset before FROTE, 2) after applying the modification strategy, and 3) after FROTE completes augmentation. In Figure~\ref{fig:fig_relabel}, additional plots for Figure 3 of the main paper are given, where the results with Splice, Nursery, Breast Cancer, Mushroom and Car datasets are included. In Figure~\ref{fig:fig_relabel}, the improvements of the $\overline{J}$ values observed after 1. modification strategy is applied, and 2. between the augmentation process and mod strategy, is displayed. Both  Figure~3 of the main paper and Figure~\ref{fig:fig_relabel} show the results with the \modrelabel{} strategy. Figure~\ref{fig:fignone1} and Figure~\ref{fig:fignone2} show the results with the \modnone{} strategy, and Figure~\ref{fig:figdrop1} and Figure~\ref{fig:figdrop2} show the results with the \moddrop{} strategy. As can be observed from the figures, the variance appears to be higher for both \modnone{} and \moddrop{} strategies, since for the former, existing contradictory instances are remained in the dataset, and for the latter, the base instances are selected through rule relaxation which increases the variety in the base instances. However, for all mod strategies, we can conclude that augmentation can improve MRA without much compromise-in some cases increase- in F1-Score.

\captionsetup[table]{skip=11pt}
\begin{figure*}
\captionsetup[subfigure]{justification=centering}
\centering
\includegraphics[width=1\textwidth, height=4cm,keepaspectratio ]{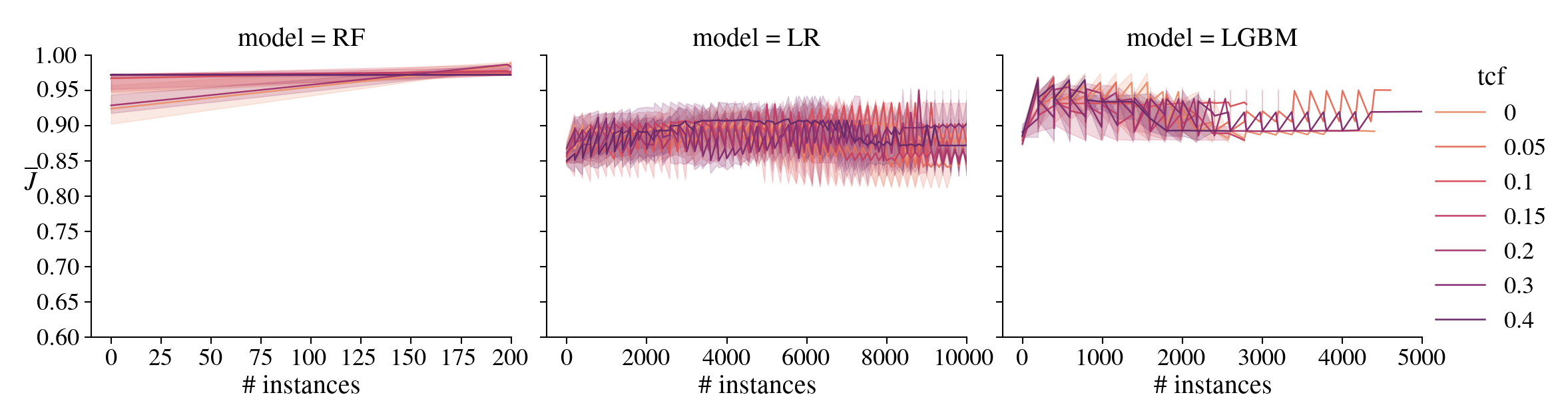}

\caption{Augmentation progress evaluated on the held-out test set for different models and $tcf$ values on the Adult dataset. The objective function $\overline{J}$ (median and 5-95 percentiles) is shown as a function of the number of instances added to the dataset during augmentation. 
Results are averaged over 90 runs, and for all runs, $|\FRS|=3$, the mod-strategy is \modrelabel, and $\BPrand$ selection is used.}
    \label{fig:augment_perf}
\end{figure*}

\textbf{Comparison with the existing work.} Additional results for the comparison experiments with~\cite{daly2021aaai} are included in Tables~\ref{table:overlay1} and~\ref{table:overlay2}. We observe from the tables that our solution performs better than a state-of-art post-processing approach, which confirms with the findings presented in the main paper. When we examine the results in Table~\ref{table:overlay2}, we see that even  \textit{Hard Constraints} has a significantly higher MRA than the \textit{Soft Constraints} for all datasets, it performs very poorly on the outside coverage population, as can be seen from the $\Delta$\textbf{F-Score} values. This demonstrates that a pure post-processing approach can suffer if the rules are deviated from the underlying model. Similar findings are observed for the \textit{Soft Constraints}, however \textit{Soft Constraints} suffers less from the deviation in the rules, since it considers models decisions after applying changes to the data instance based on the rules learnt so far.

\textbf{Augmentation progress.}
In Figure~\ref{fig:augment_perf}, we evaluate $\overline{J}$ on the held-out test set for intermediate models trained on $D'$ (i.e.~augmented training dataset at the end of each iteration) as a function of the number of synthetic instances added, to illustrate how these change for different models and $tcf$ values. 
For all models, $\overline{J}$ improves more quickly for lower training coverage. RF needs fewer instances to reach $\overline{J}=1$ in comparison with LR and LGBM. This again suggests that non-linear models like RF may require less data to edit than linear models. 

\begin{figure*} [th]
    \centering
        \includegraphics[width=11cm,height=8cm,keepaspectratio]{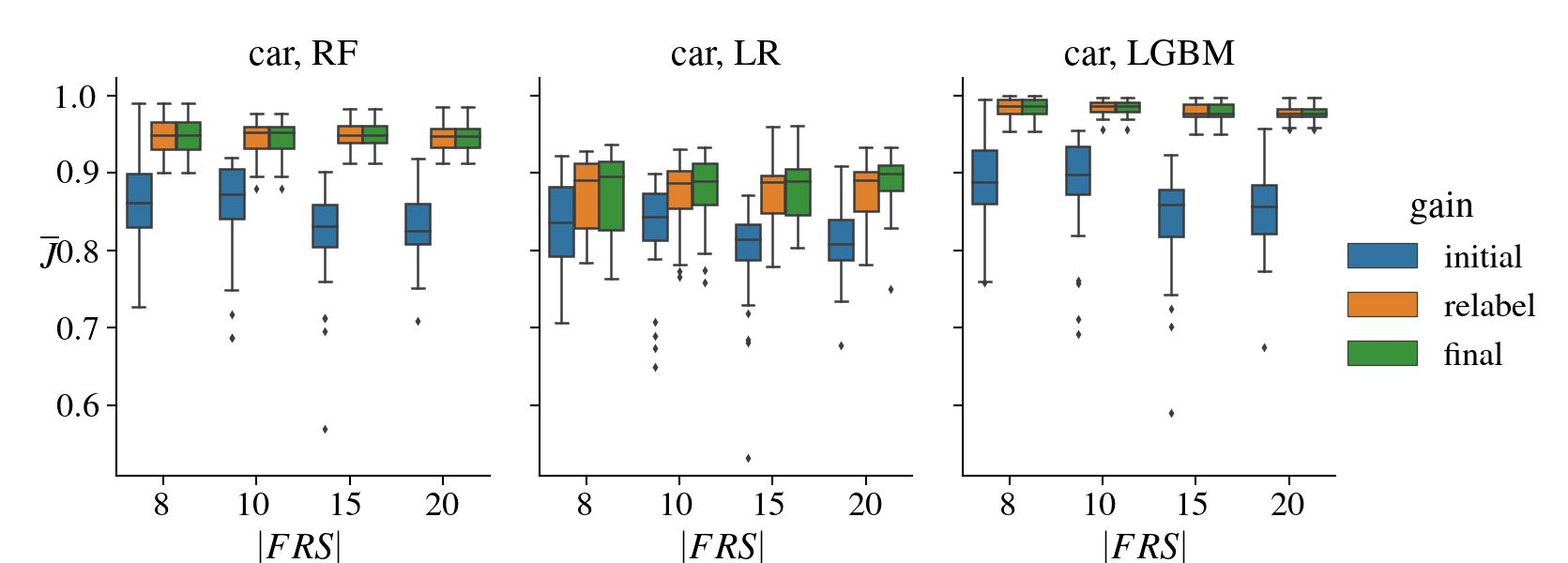}
        \includegraphics[width=11cm,height=8cm,keepaspectratio]{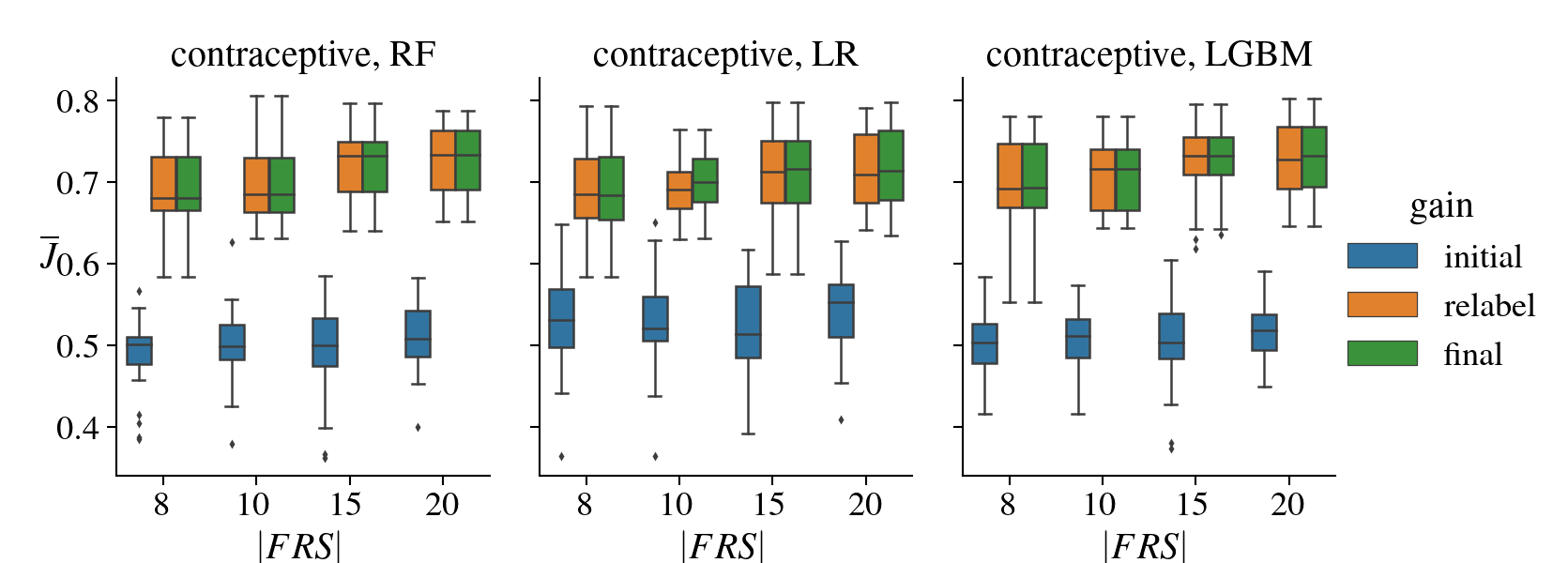}
        \includegraphics[width=11cm,height=8cm,keepaspectratio]{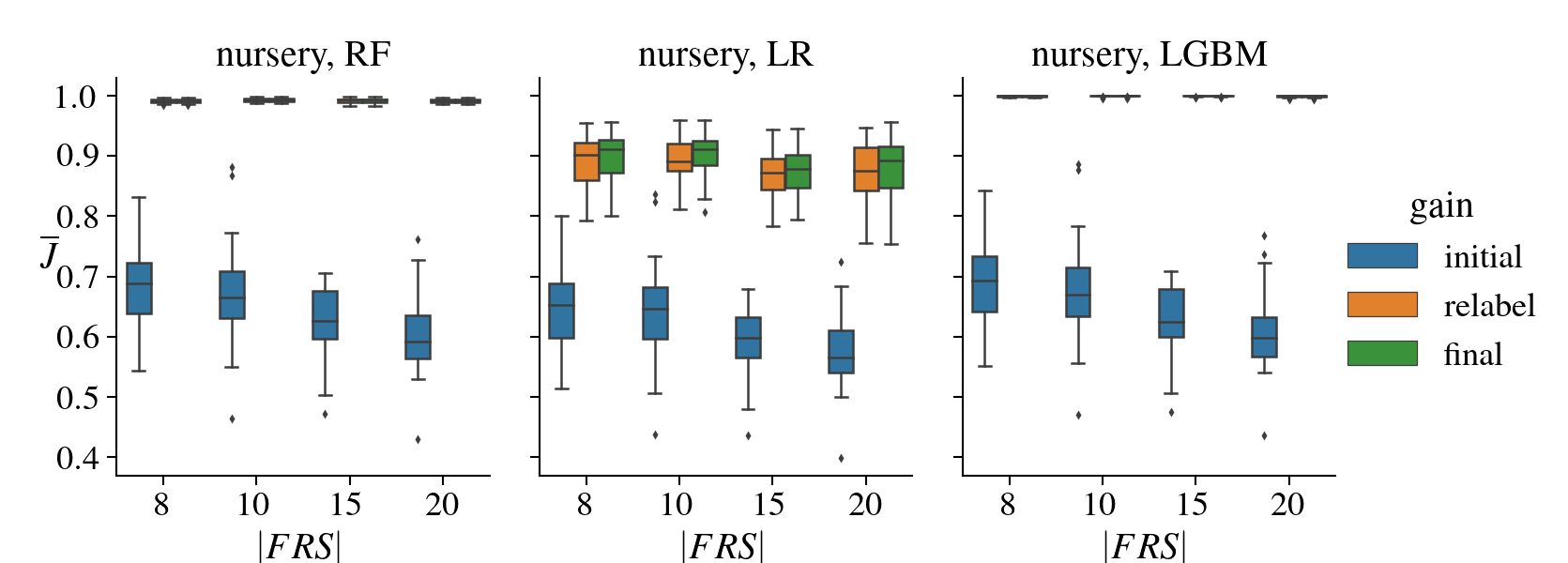}
        \includegraphics[width=11cm,height=8cm,keepaspectratio]{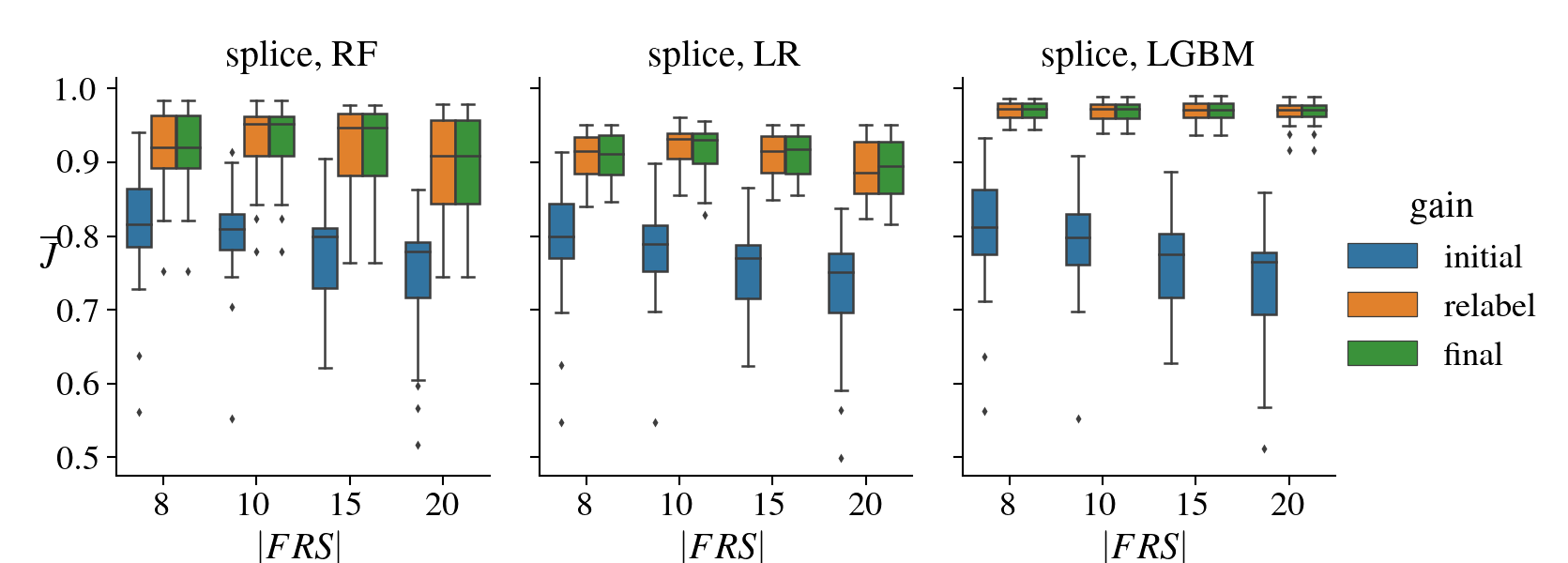}
     \caption{Additional plots for Figure~3 in the main paper. Effect of feedback rule set size for the Car, Contraceptive, Nursery and Splice datasets are given using the \BPrand{} selection strategy. The same comparison as in Figure 1 is shown between initial (before FROTE), after relabel, and final (after augmentation). Each box and whiskers is computed from 20 runs with $tcf = 0.2$, $\alpha = 0.8$, $k=5$.}
    \label{fig:fig_number_of_rules}
\end{figure*}

\textbf{Number of feedback rules.} Additional plots for displaying the effect of number of rules on the performance of the solution are given in Figure~\ref{fig:fig_number_of_rules}. For all datasets, we experimented with |$\FRS$|=\{8,10,15,20\}, however for some datasets, for |$\FRS$|=15 and |$\FRS$|=20, no such conflict-free $\FRS$ can be found out of $500$ rules. Therefore, we included the results for the experiments for which a conflict-free rule set with the experimented size can be formed. 

As it is observed from the results, FROTE improves
$J$ both after the relabel modification strategy and after the data augmentation. Overall, results demonstrate the efficacy of the approach even with larger rule sets.

\textbf{Base  instance selection  strategy.} Performance of the two base instance selection strategies, \BPIP{} and \BPrand{} were compared in the main paper using the improvement in $\overline{J}$. In Table~\ref{table:IPvsAllTMain}, we have included the number of instances added to achieve those improvements. In Table~\ref{table:IPvsAllT_mra}, improvements in MRA and F-Score are reported separately. We observe that the improvement in $\overline{J}$ is highly dominated by the improvement in MRA. 

\begin{table*}[th]
    \footnotesize
    \centering
    \captionsetup{font=small}
    \caption{Experiments with \textit{\BPIP} and \textit{\BPrand} selection strategies for all datasets and models. Number of instances are included as an additional column to Table~2 of the main paper. $\Delta\#Ins/|D|$ is the number of instances added (as a fraction of the dataset size) that leads to the reported improvements. Means and standard deviations are computed from all runs performed with a given dataset and model.}
    \vspace{2mm}
    \begin{tabular}{p{1.35cm}| p{0.4cm}| p{1.9cm}| p{1.9cm}| p{2cm} | p{1.75cm}} 
    \hline 
       \multicolumn{1}{l|}{\textbf{Dataset}} & \multicolumn{1}{|l|}{\textbf{Model}} & \multicolumn{1}{|l|}{\textbf{$\Delta \overline{J}$ (\BPrand)}} &
       \multicolumn{1}{|l|}{\textbf{$\Delta \overline{J}$ (\BPIP)}} & \multicolumn{1}{|l|}{\textbf{$\Delta$\#Ins$/|D|$  (\BPrand)}} &
       \multicolumn{1}{|l}{\textbf{$\Delta$\#Ins$/|D|$ (\BPIP)}}\\\hline

\textbf{\shortstack[l]{B.Cancer}}
&RF & $0.000\pm0.003 $ & $0.001\pm0.006 $& $0.011\pm0.016$ & $0.015\pm0.042$\\
&LR & $0.006\pm0.022 $ & $0.006\pm0.026 $& $0.298\pm0.326$ & $0.199\pm0.266$\\
&LGBM & $0.001\pm0.008 $ & $0.002\pm0.010 $& $0.011\pm0.016$ & $0.014\pm0.036$\\
\hline
\textbf{\shortstack[l]{Car}}
&RF & $0.005\pm0.020 $ & $0.006\pm0.020 $& $0.001\pm0.003$ & $0.001\pm0.004$\\
&LR & $0.022\pm0.034 $ & $0.020\pm0.029 $& $0.227\pm0.225$ & $0.113\pm0.183$\\
&LGBM & $0.008\pm0.033 $ & $0.008\pm0.027 $& $0.001\pm0.003$ & $0.001\pm0.004$\\
\hline
\textbf{\shortstack[l]{Mushroom}}
&RF & $0.001\pm0.017 $ & $0.004\pm0.034 $& $0.001\pm0.002$ & $0.001\pm0.002$\\
&LR & $0.005\pm0.023 $ & $0.011\pm0.049 $& $0.036\pm0.136$ & $0.016\pm0.064$\\
&LGBM & $0.004\pm0.037 $ & $0.006\pm0.041 $& $0.001\pm0.002$ & $0.001\pm0.002$\\
\hline
\textbf{\shortstack[l]{Adult}}
&RF & $0.003\pm0.014 $ & $0.003\pm0.011 $& $0.005\pm0.047$ & $0.004\pm0.009$\\
&LR & $0.008\pm0.023 $ & $0.004\pm0.012 $& $0.356\pm0.507$ & $0.185\pm0.352$\\
&LGBM & $0.004\pm0.015 $ & $0.003\pm0.011 $& $0.059\pm0.096$ & $0.046\pm0.083$\\
\hline
\textbf{\shortstack[l]{Wine}}
&RF & $0.001\pm0.007 $ & $0.001\pm0.007 $& $0.004\pm0.033$ & $0.003\pm0.016$\\
&LR & $0.056\pm0.096 $ & $0.055\pm0.094 $& $0.136\pm0.135$ & $0.096\pm0.098$\\
&LGBM & $0.003\pm0.015 $ & $0.003\pm0.01 $& $0.002\pm0.009$ & $0.003\pm0.009$\\
\hline
\textbf{\shortstack[l]{Contracep.}}
&RF & $0.032\pm0.081 $ & $0.038\pm0.085 $& $0.000\pm0.001$ & $0.001\pm0.001$\\
&LR & $0.041\pm0.099 $ & $0.051\pm0.102 $& $0.011\pm0.019$ & $0.008\pm0.013$\\
&LGBM & $0.027\pm0.066 $ & $0.026\pm0.057 $& $0.001\pm0.003$ & $0.001\pm0.003$\\
\hline
\textbf{\shortstack[l]{Nursery}}
&RF & $0.031\pm0.099 $ & $0.023\pm0.076 $& $0.001\pm0.003$ & $0.001\pm0.002$\\
&LR & $0.043\pm0.088 $ & $0.029\pm0.069 $& $0.144\pm0.162$ & $0.031\pm0.044$\\
&LGBM & $0.035\pm0.108 $ & $0.030\pm0.096 $& $0.001\pm0.003$ & $0.001\pm0.002$\\
\hline
\textbf{\shortstack[l]{Splice}}
&RF & $0.003\pm0.017 $ & $0.002\pm0.012 $& $0.009\pm0.047$ & $0.008\pm0.044$\\
&LR & $0.011\pm0.031 $ & $0.007\pm0.018 $& $0.091\pm0.116$ & $0.046\pm0.079$\\
&LGBM & $0.014\pm0.049 $ & $0.009\pm0.037 $& $0.009\pm0.047$ & $0.007\pm0.040$\\
\hline

    \end{tabular}
    \label{table:IPvsAllTMain}
\end{table*}

\begin{table*}[th]
    \footnotesize
    \centering
    \captionsetup{font=small}
    \caption{MRA and F-Score reported separately for the results in Table~1 of the main paper. Same with Table 1, results are reported for \textit{\BPIP} and \textit{\BPrand} selection strategies.$\Delta$MRA and $\Delta$F-Score represent the improvement in the corresponding metrics ($mean\pm std$). Means and standard deviations are computed from all runs performed with a given dataset and model.}
    \vspace{2mm}

    \begin{tabular}{p{2cm}| p{0.4cm}| p{1.9cm}| p{1.9cm}| p{2cm} | p{2.2cm}} 
    \hline 
       \multicolumn{1}{l|}{\textbf{Dataset}} & \multicolumn{1}{|l|}{\textbf{Model}} & \multicolumn{1}{|l|}{$\Delta$ \textbf{MRA (\BPIP)}} &
       \multicolumn{1}{|l|}{$\Delta$\textbf{MRA (\BPrand)}} & \multicolumn{1}{|l|}{$\Delta$\textbf{F-Score  (\BPIP)}} &
       \multicolumn{1}{|l}{$\Delta$\textbf{F-Score (\BPrand)}}\\\hline

\textbf{\shortstack[l]{Breastcancer}}
&RF &$0.003\pm0.042$&$0.002\pm0.038$
&$0.000\pm0.005$&$0.000\pm0.003$\\
&LR &$0.047\pm0.116$&$0.039\pm0.102$
&$\mathllap-0.006\pm0.014$&$\mathllap-0.006\pm0.015$\\
&LGBM &$0.013\pm0.092$&$0.014\pm0.098$
&$0.000\pm0.006$&$0.000\pm0.005$\\
\hline

\textbf{\shortstack[l]{Car}}
&RF &$0.018\pm0.063$&$0.015\pm0.069$
&$0.000\pm0.003$&$0.000\pm0.003$\\
&LR &$0.096\pm0.112$&$0.109\pm0.135$
&$\mathllap-0.020\pm0.028$&$\mathllap-0.026\pm0.031$\\
&LGBM &$0.024\pm0.083$&$0.024\pm0.099$
&$0.000\pm0.002$&$0.000\pm0.002$\\
\hline

\textbf{\shortstack[l]{Mushroom}}
&RF &$0.009\pm0.081$&$0.002\pm0.027$
&$\mathllap-0.000\pm0.000$&$\mathllap-0.000\pm0.000$\\
&LR &$0.045\pm0.158$&$0.024\pm0.111$
&$\mathllap-0.000\pm0.001$&$\mathllap-0.000\pm0.001$\\
&LGBM &$0.024\pm0.141$&$0.018\pm0.128$
&$\mathllap-0.000\pm0.000$&$\mathllap-0.000\pm0.000$\\
\hline
\textbf{\shortstack[l]{Adult}}
&RF &$0.011\pm0.053$&$0.012\pm0.073$
&$\mathllap-0.000\pm0.001$&$-0.0\pm0.001$\\
&LR &$0.072\pm0.170$&$0.075\pm0.192$
&$\mathllap-0.003\pm0.005$&$\mathllap-0.003\pm0.007$\\
&LGBM &$0.026\pm0.108$&$0.026\pm0.117$
&$0.000\pm0.001$&$0.000\pm0.001$\\
\hline
\textbf{\shortstack[l]{Wine}}
&RF &$0.018\pm0.096$&$0.020\pm0.107$
&$\mathllap-0.001\pm0.005$&$\mathllap-0.000\pm0.005$\\
&LR &$0.360\pm0.306$&$0.354\pm0.309$
&$\mathllap-0.020\pm0.023$&$\mathllap-0.023\pm0.026$\\
&LGBM &$0.043\pm0.169$&$0.037\pm0.155$
&$0.001\pm0.005$&$0.001\pm0.008$\\
\hline
\textbf{\shortstack[l]{Contraceptive}}
&RF &$0.070\pm0.151$&$0.059\pm0.144$
&$\mathllap-0.000\pm0.009$&$\mathllap-0.000\pm0.007$\\
&LR &$0.115\pm0.214$&$0.095\pm0.203$
&$\mathllap-0.007\pm0.019$&$\mathllap-0.010\pm0.025$\\
&LGBM &$0.048\pm0.104$&$0.049\pm0.119$
&$\mathllap-0.001\pm0.010$&$\mathllap-0.000\pm0.009$\\
\hline

\textbf{\shortstack[l]{Nursery}}
&RF &$0.059\pm0.192$&$0.074\pm0.226$
&$\mathllap-0.000\pm0.001$&$\mathllap-0.000\pm0.001$\\
&LR &$0.097\pm0.217$&$0.131\pm0.240$
&$\mathllap-0.002\pm0.004$&$\mathllap-0.008\pm0.013$\\
&LGBM &$0.073\pm0.227$&$0.082\pm0.242$
&$\mathllap-0.000\pm0.000$&$\mathllap-0.000\pm0.000$\\
\hline

\textbf{\shortstack[l]{Splice}}
&RF &$0.006\pm0.026$&$0.009\pm0.036$
&$\mathllap-0.001\pm0.004$&$\mathllap-0.001\pm0.004$\\
&LR &$0.025\pm0.046$&$0.035\pm0.071$
&$\mathllap-0.002\pm0.006$&$\mathllap-0.004\pm0.010$\\
&LGBM &$0.022\pm0.098$&$0.032\pm0.116$
&$0.000\pm0.002$&$0.000\pm0.002$\\
\hline

    \end{tabular}
    \label{table:IPvsAllT_mra}
\end{table*}

\paragraph{Probabilistic rules.} In this experiment, we consider probabilistic rules, where the label distribution $\pi$ is not just a Kronecker delta for one of the classes. The experiment provides a brief demonstration of the ability of probabilistic rules to represent uncertainty and mitigate the effect of an over-confident expert rule. We consider an extreme case of this where the expert provides a single feedback rule, but the test distribution remains the same as the training distribution, i.e., the expert is wrong and the rule does not take effect. (We use only a single feedback rule to try to isolate the effect of having a probabilistic rule and avoid interactions among rules.) We also set $tcf=0$ (so \modrelabel{} and \moddrop{} initializations are not applicable).

We run FROTE with the following label distribution $\pi$ for instances generated under the rule: With probability $p$, the label is equal to the class $c$ specified by the feedback rule. With probability $1-p$, it is equal to the label of the corresponding base instance, except when that label is $c$, in which case the label of the generated instance is chosen uniformly at random from classes other than $c$. Thus overall, the labels of generated instances are equal to $c$ with probability $p$, and otherwise they approximately follow the distribution of the training data (as represented by the base instances) restricted to classes other than $c$. The case $p = 1$ is the deterministic case used in the other experiments. With $p < 1$, the user of FROTE can express less than full confidence in the expert rule and rely more on the existing training data.

Table~\ref{table:probrules} shows the MRA and $\overline{J}$ improvements for different probabilities $p$. In this case, since the feedback rule is not in effect for test data, MRA just measures agreement with respect to labels following the original distribution $p_{X,Y}$, within the coverage of the rule. The MRA column shows that setting $p = 1.0$, i.e., completely following the expert rule, does not give as good a performance as setting $p$ to a lower, less confident value. This pattern however is not as clear looking at the $\overline{J}$ column. In reality, the best value of $p$ is not known a priori as it depends on the exact extent to which the test data (in this case, the distribution $p_{X,Y}$) conforms to the expert rule. Nevertheless, Table~\ref{table:probrules} suggests that there is a benefit to using a probabilistic rule with $p < 1$ if there is reason to be less confident in the validity of the feedback rules.

\begin{table}[ht]
    \small
    \centering
    \caption{Experiments with probabilistic rules. Means and standard deviations computed from $50$ runs for LR model and for the given datasets. For each run, $|FRS| = 1$, and $tcf=0$. \BPrand{} selection strategy is utilized during the experiments.}
    \vspace{2mm}
    \begin{tabular}{@{}llcc@{}}
        \toprule
        Dataset & Probability & $\Delta {mra}$ & $\Delta \overline{J}$ \\
        \midrule
        Mushroom & $p=0.4$ & $0.206\pm0344$ & $0.007\pm0.012 $\\
        &$p=0.6$ & $0.242\pm0.386$ & $0.009\pm0.014$\\
        &$p=0.8$ & $0.249\pm0.390$ & $0.009\pm0.014$\\
        &$p=1.0$ & $0.173\pm0.296$ & $0.006\pm0.011$\\\hline
        Wine & $p=0.4$ & $0.416\pm0.305$ & $\mathllap\shortminus0.011\pm0.024$\\
        &$p=0.6$ & $0.448\pm0.317$ & $\mathllap\shortminus0.010\pm0.021$\\
        &$p=0.8$ & $0.423\pm0.348$ & $\mathllap\shortminus0.011\pm0.020$\\
        &$p=1.0$ & $0.338\pm0.327$ & $\mathllap\shortminus0.008\pm0.016$\\\hline
        B.~Cancer & $p=0.4$ & $0.005\pm0.015$ & $0.003\pm0.007$\\
        &$p=0.6$ & $0.005\pm0.015$ & $0.002\pm0.006$\\
        &$p=0.8$ & $0.007\pm0.015$ & $0.002\pm0.007$\\
        &$p=1.0$ & $0.005\pm0.015$ & $0.003\pm0.006$\\
        \bottomrule
    \end{tabular}
    
    \label{table:probrules}
\end{table}

\begin{table}[t]
    \small
    \centering
    \caption{Comparison with Overlay-Soft (soft constraints) and Overlay-Hard (hard constraints) of \citet{daly2021aaai} on Adult dataset. Means and standard deviations computed from 50 runs.}
    \vspace{2mm}
    \begin{tabular}{@{}p{0.9cm}p{0.7cm}p{1.8cm}p{1.8cm}p{1.8cm}@{}}
        \toprule
        Dataset & Model & \multicolumn{3}{c}{$\Delta \overline{J}$}\\
        \cmidrule(lr){3-5}
         & & Overlay-Soft & Overlay-Hard & FROTE\\
        \midrule
        Adult & LR & $\mathllap\shortminus0.015\pm0.034$ & $\mathllap\shortminus0.107\pm0.111$ & $0.025\pm0.039$\\
        & RF & $0.114\pm0.013$ & $\mathllap\shortminus0.121\pm0.019$ & $0.036\pm0.039$\\
        & LGBM & $0.102\pm0.021$ & $\mathllap\shortminus0.018\pm0.180$ & $0.240\pm0.043$\\
        \bottomrule
    \end{tabular}

    \label{table:overlay1}
\end{table}

\begin{table*}
\captionsetup{font=small}
    \caption{Experiments with \textit{Overlay}~\cite{daly2021aaai}. Overlay-Hard and Overlay-Soft refers to the Hard Constraints and Soft Constraints approaches of Overlay. The comparison is shown for different ML models and the Breast Cancer, Mushroom and Adult datasets. \BPrand{} selection strategy is used for FROTE. Means and standard deviations are computed from 50 runs, where for each run a different set of $3$ rules are used. }
    \vspace{2mm}
  \centering
  \scriptsize
\begin{tabular}{|l|l|l|l|l|l|l|}
    \hline
    \multicolumn{1}{|c|}{\textbf{Model}} & \multicolumn{3}{c|}{$\Delta$\textbf{MRA}} & \multicolumn{3}{c|}{$\Delta$\textbf{F-Score}} \\
    \hline
    \multicolumn{7}{|c|}{\textbf{B. Cancer}}\\
   \hline
      & Overlay-Soft & Overlay-Hard  & FROTE& Overlay-Soft & Overlay-Hard & FROTE\\
    \hline
   
    LR & $0.008\pm0.021$ & $0.021\pm0.232$ &$ 0.080\pm0.168$ & $\mathllap-0.012\pm0.058$ & $\mathllap-0.313\pm 0.248$ & $\mathllap-0.014\pm0.022$\\\hline
    RF & $0.005\pm0.015$ & $0.071\pm0.224$ &$0.097\pm0.194$ & $0.000\pm0.000$ & $\mathllap-0.299\pm0.238$ & $\mathllap-0.002\pm0.007$ \\
    \hline
    LGBM & $0.016\pm0.032$ & $0.072\pm0.218$ & $0.880\pm0.238$ & $\mathllap-0.000\pm0.001$ & $\mathllap-0.272\pm0.198$& $\mathllap-0.001\pm0.009$ \\\hline
    \multicolumn{7}{|c|}{\textbf{Mushroom}}\\
     \hline
     & Overlay-Soft & Overlay-Hard & FROTE
    & Overlay-Soft & Overlay-Hard & FROTE \\
   \hline
    LR & $0.046\pm0.091$ & $0.202\pm0.34$ & $0.049\pm0.033$ & $\mathllap-0.001\pm0.004$ & $\mathllap-0.168\pm0.223 $ & $\mathllap-0.000\pm0.001$ \\\hline
    RF & $0.021\pm0.114$ & $0.205\pm0.34$ & $0.040\pm0.032$ & $0.000\pm0.000$ & $\mathllap-0.166\pm0.220$ &$0.000\pm0.000$ \\\hline
    LGBM & $0.155\pm0.302$ & $0.208\pm0.34$ & $0.049\pm0.033$ & $\mathllap-0.023\pm0.093$ & $\mathllap-0.163\pm0.218$ & $0.000\pm0.000$ \\
    \hline
\end{tabular}
\label{table:overlay2}
\end{table*}


\end{document}